\documentclass{article}

\usepackage{PRIMEarxiv}
\usepackage[utf8]{inputenc} % allow utf-8 input
\usepackage[T1]{fontenc}    % use 8-bit T1 fonts
\usepackage{hyperref}       % hyperlinks
\usepackage{url}            % simple URL typesetting
\usepackage{booktabs}       % professional-quality tables
\usepackage{amsfonts}       % blackboard math symbols
\usepackage{nicefrac}       % compact symbols for 1/2, etc.
\usepackage{microtype}      % microtypography
\usepackage{lipsum}
\usepackage{fancyhdr}       % header
\usepackage{graphicx}       % graphics

\usepackage{subfig}
\usepackage{empheq}
\usepackage{longtable}
\usepackage{amsmath}
\usepackage{amsthm}
\usepackage{float}
\usepackage{tabularx}
\usepackage{booktabs}
\usepackage{url}
\newcommand{\reffig}[1]{Fig.~\ref{#1}}
\newcommand{\refeqn}[1]{Equation~(\ref{#1})}
\newcommand{\reftab}[1]{Table~\ref{#1}}

\newcommand{\refsec}[1]{Section~\ref{#1}}

\newcommand{\expnum}[2]{{#1}\mathrm{E}{#2}}
\newcommand{\munit}[1]{\mathrm{#1}}
\newcommand{\vekt}[1]{\boldsymbol{#1}}
\newcommand{\matr}[1]{\boldsymbol{#1}}

\newcommand{\pderiv}[2]{\frac{\partial #1}{\partial #2}}

\newcommand{\bs}[1]{\boldsymbol{#1}}
\newcommand{\vel}{\vekt{v}}
\newcommand{\xnew}{\vekt{x}}
\newcommand{\xa}{\vekt{x}}

\newcommand{\nab}[1]{\vekt{\nabla}_{#1}}
\newcommand{\nabx}{\nab{\bs{x}}}
\newcommand{\nabxd}{\nab{\bs{x}^*}}
\newcommand{\lap}[1]{\vekt{\nabla}_{#1}^2}
\newcommand{\lapx}{\lap{\bs{x}}}

\newcommand{\reynolds}{\textrm{Re}}
\newcommand{\Pas}{\,Pa\,s}

\title{Investigation of Physics-Informed Deep Learning for the Prediction of Parametric, Three-Dimensional Flow Based on Boundary Data}
\author{
	Philip Heger\\
	Chair for Computational Analysis of Technical Systems/Thermische Betriebssicherheit (EG-623)\\
	Chair for Computational Analysis of Technical Systems/BMW AG\\
	Schinkelstr. 2, 52062, Aachen/Petuelring 130, 80809, Munich\\
	\texttt{philip.heger@rwth-aachen.de} \\
	\AND
	Markus Full\\
	Thermische Betriebssicherheit (EG-623)\\
	BMW AG\\
	Petuelring 130, 80809, Munich\\
	\texttt{markus.full@bmw.de} \\
	\AND
	Daniel Hilger, Norbert Hosters\\
	Chair for Computational Analysis of Technical Systems\\
	RWTH Aachen University\\
	Schinkelstr. 2, 52062, Aachen\\
	\texttt{\{hilger, hosters\}@cats.rwth-aachen.de} \\
	\textit{www.cats.rwth-aachen.de}
	
}

\theoremstyle{remark}

\begin{document}
	\maketitle
	
\begin{abstract}
  		The placement of temperature sensitive and safety-critical components is crucial in the automotive industry.
	  	It is therefore inevitable, even at the design stage of new vehicles that these components are assessed for potential safety issues.
  		However, with increasing number of design proposals, risk assessment quickly becomes expensive.
	  	We therefore present a parameterized surrogate model for the prediction of three-dimensional flow fields in aerothermal vehicle simulations. The proposed physics-informed neural network (PINN) design is aimed at learning families of flow solutions according to a geometric variation. In scope of this work, we could show that our nondimensional, multivariate scheme can be efficiently trained to predict the velocity and pressure distribution for different design scenarios and geometric scales. The proposed algorithm is based on a parametric minibatch training which enables the utilization of large datasets necessary for the three-dimensional flow modeling. Further, we introduce a continuous resampling algorithm that allows to operate on one static dataset. Every feature of our methodology is tested individually and verified against conventional CFD simulations. Finally, we apply our proposed method in context of an exemplary real-world automotive application.
\end{abstract}

	\section{Introduction}\label{sec:introduction}
	In automotive development it is essential to ensure that vehicles can be operated safely in all 
	expected 
	scenarios. This particularly applies to the thermal safety of temperature-sensitive or safety-critical 
	components.
	Temperature reduction on these components is mainly enforced through convective heat transfer by the surrounding air\cite{Hahndel2014,Saad2021}. 
	One established method to determine suitable component geometries is the simulation of the fluid flow and heat transfer inside the vehicles.
	These simulations, however, contain complex models with large memory and computation time 
	requirements to represent a physically plausible scenario.
	Resource heavy computations are unsuitable in the early design stage where many geometry and 
	component parameters are still variable or possibly unknown.
	Every new design suggestion requires an estimation of the expected temperature on the critical components.
	Therefore, there is a clear incentive to replace the complex simulations with reduced order models (ROM) that predict the temperatures at critical components under varying design parameters sufficiently accurate \cite{audms}.\\
	
	Reducing the computational costs and simulation times has always been of interest to the computational mechanics community.
	Hence, there exist already a multitude of different ROM strategies.
	One way to categorize ROM is to group them into hierarchical, projection based and data-driven 
	approaches \cite{EldredDunlavy2006,BennerGugercinWillcox2015}.
	Hierarchical ROM refers to models that simplify the underlying physics \cite{RozzaEtAl2018}, whereas in projection based methods the full-order operators are projected onto a reduced basis.
	The reduced basis allows for fast evaluation of the system response as it has been shown in examples presented in \cite{QuarteroniManzoniNegri2015, HesthavenRozzaStamm2016, GuoHesthave2018, 
		KeyEtAl2021}.
	Although, projection-based ROM methods already provide accurate results, the data-driven ROM approach in particular is of special interest in context of automotive development.
	This is because data-driven ROM methods, by relying on empirical data, do not simplify any physics and do not require access to the source code of numerical solvers.
	The terminology of data-driven ROM once again contains a multitude of different strategies. 
	The number of methods especially increased with the latest advances in machine learning (ML).
	In combination with ML many novel ROM approaches were created, e.g. regression based methods as presented in \cite{Waxenegger_Wilfing_2020,dlr130206,10.1007/978-3-030-22808-8_24, BerzinsEtAl2020,HesthavenUbbiali2018}, convolutional autoencoders \cite{MaulikLuschBalaprakash2021} or physics-informed deep learning  \cite{raissi2018hidden,RAISSI2019686,10.1145/3394486.3403198}, to only name a few.
	With increasing computational resources and data availability ML offers a perspective to approximate complex relations while allowing for repeated and fast evaluations once a model is trained.
	In scope of this work we want to focus on the method of physics-informed neural networks (PINN), which were initially introduced by Raissi et al. to handle sparse training data by leveraging formerly unused knowledge about the system behavior \cite{RAISSI2019686}. \\

	In case of PINN the additional knowledge is provided by embedding governing partial differential equations (PDE) of the system into the learning process of a black box surrogate model.
	This constrains the trained parameters to a physically plausible solution space.
	In \cite{RAISSI2019686} and \cite{HAGHIGHAT2021113552} the PINN is used to solve different sets of nonlinear PDE in one spatial dimension and time, using continuous as well as discontinuous time models. 
	It is also shown that the PINN can be used to discover unknown system parameters from measurement data.
	Arthurs et al. extend the forward prediction to a more complex set of nonlinear PDE in the form of the incompressible Navier-Stokes equations in two dimensions \cite{arthurs2020active}. 
	The flow field inside a set of tubular domains is predicted based on sparse volume data. 
	Furthermore, the interpolation capability of the PINN is used to directly train a continuous, parametric solution for a variable geometry parameter using an active learning approach. 
	%In \cite{raissi2018hidden} Raissi et al. show that using a physics-informed architecture, a flow 
	%field can be recovered from the knowledge about a passive scaler like concentration data.
	In \cite{SUN2020112732} and \cite{RAO2020207} PINN designs are proposed for the solution of flow equations without any volume data. 
	Rao et al. introduce a multivariate formulation of the two-dimensional Navier-Stokes equations to significantly improve training performance \cite{RAO2020207}.
	Since PINN were first introduced, the concept was continuously extended to incorporate models that are already state of the art in classical computational fluid dynamics, as for example turbulence modeling \cite{ChengEtAl2021}.
	At last it is to mention the idea of including a parameterization of the PDEs into the structure of PINN as presented in \cite{demoStrazulloRozza2021}.
	Though Demo et Al. mention that the parameterization concept could be applied to geometry variations, their work is limited to variations of physical features.\\

	This work aims to develop a PINN design for the prediction of three-dimensional velocity and pressure distributions solely from BC data.
	Most existing research has limited its scope to either two spatial dimensions, the use of volume data or strongly simplified geometries. 
	To investigate the viability of the PINN architecture in the context of modeling vehicle components, it is necessary to handle arbitrary geometries and variable scales of boundary conditions. 
	It is also examined whether the PINN can learn a continuous family of solutions according to a variable geometric parameter. 
	An online resampling algorithm is presented to handle the distribution of training data without increasing the dataset size. 
	For the assumption of incompressible fluids the solution is decoupled from the conservation of energy and thus thermal conditions \cite{nla.cat-vn1911901}.
	Reliable and physically plausible approximations enable the utilization of such flow fields for coupled simulations with established thermal solvers. 
	To the best knowledge of the authors this is one of the first attempts in literature to extend the PINN approach to the forward prediction of three-dimensional, parametric flow solutions without any volume data.
	The implemented code can be found via \cite{Software2022} and the corresponding datasets of the presented work is accessible via \cite{Dataset2022}.
	
	\section{Governing Equations}
	\label{sec:methods}
	For a steady problem the motion and temperature of an incompressible fluid are governed by the Navier-Stokes equations depicted in \refeqn{eq:nav-incomp-v-p-steady} \cite{rg_segregated_energy_incompressible}. 
	\refeqn{eq:nav-incomp-v-p-stead_mass} and \refeqn{eq:nav-incomp-v-p-stead_mom} are the conservation of mass and the conservation of momentum respectively \cite{fef_book_ch1}. 
	The equations are written using the primitive variables density $\rho$, pressure $p$ and velocity $\vel \in \mathbb{R}^{n_{sd}}$.
	$\matr{\sigma}$ denotes the symmetric Cauchy stress tensor.
	\refeqn{eq:nav-incomp-v-p-stead_ene} describes the energy balance represented as the static temperature equation \cite{visc_incomp_fluid,starccm}. 
	$Q$ is the external volumetric heat source.
	The mass equation corresponds to a divergence free velocity field. 
	Due to the absence of an equation of state in incompressible flows, the temperature equation is decoupled from the mass- and momentum balance.
	It can therefore be neglected for the solution of flow problems. \cite{nla.cat-vn1911901}
	\begin{subequations}
		\label{eq:nav-incomp-v-p-steady}
		\begin{equation}
			\label{eq:nav-incomp-v-p-stead_mass}
			\nabx \cdot \vel = 0 \,,
		\end{equation}
		\begin{equation}
			\label{eq:nav-incomp-v-p-stead_mom}
			\big(\vel \cdot \nabx\big)\vel\ = \nabx \cdot \matr{\sigma} \,,
		\end{equation}
		\begin{equation}
			\label{eq:nav-incomp-v-p-stead_ene}
			\rho c \vel \cdot \nabx T = \lambda \lapx T + Q \, .
		\end{equation}
	\end{subequations}
	Using the divergence free condition, the Cauchy stress reduces to \refeqn{eq:stokes-law}.
	This relation is called Stokes' law \cite{fef_book_ch1}.
	$\mu$ here denotes the fluid dynamic viscosity.
	\begin{equation}
		\label{eq:stokes-law}
		\matr{\sigma}(\vel,p) = -p\vekt{I} + \mu \bigg(\nabx\vel + \big(\nabx\vel\big)^T\bigg) \,.
	\end{equation}
	To form a well-posed boundary value problem the system of equations is closed by suitable boundary conditions as shown in \refeqn{eq:initialBDcond}.
	On the Dirichlet portion of the boundary $\Gamma_{D}$ the velocity is prescribed.
	On the Neumann portion of the boundary $\Gamma_{N}$ the traction $\vekt{t}$ is given \cite{fef_book_ch1}.
	\begin{subequations}
		\label{eq:initialBDcond}
		\begin{align}
			\vel &= \vel_D  \,, &\textrm{ on } \Gamma_{D} \,, \\
			\bs{n} \cdot \matr{\sigma}&= \vekt{t}  \,, &\textrm{ on } \Gamma_{N}  \,.
		\end{align}
	\end{subequations}
	A challenge in the handling of different BC are the potentially wide ranges of physical quantities such as position, velocity as well as pressure.
	As these quantities are connected through physical units and must fulfill the imposed relations, they cannot be scaled independently using normalization approaches established in deep learning.
	Normalization of data, however, is one of the key factors to good convergence behavior of deep learning algorithms \cite{pytorch2020}.
	To ensure that this condition is met for all quantities as well as their gradients, the use of a dimensionless formulation is proposed for data normalization.
	To de-dimensionalize the Navier-Stokes equations a reference length scale $L_{ref}$ and a 
	characteristic reference velocity $V_{ref}$ are defined \cite{solid_book.ch1}.
	$\xnew^*$, $\vel^*$ and $p^*$ denote the nondimensional position, velocity and pressure respectively and are defined as in \refeqn{eq:nondimQuantities} \cite{solid_book.ch1}. 
	The reference parameters $L_{ref}$ and $V_{ref}$ are used to scale the data based on the 
	expected value ranges prior to training. 
	\begin{subequations}
		\label{eq:nondimQuantities}
		\begin{align}
			\xnew^* &= \frac{1}{L_{ref}} \xnew  \,,\\
			\vel^* &= \frac{1}{V_{ref}} \vel  \,,\\
			p^* &= \frac{1}{\rho V_{ref}^2} p  \,.
		\end{align}
	\end{subequations} 
	Inserting these definitions into \refeqn{eq:nav-incomp-v-p-steady} and neglecting the energy equation for flow prediction yields the dimensionless form of the Navier-Stokes equations \cite{fef_book_ch1}:
	\begin{equation}
		\label{eq:cauchyDim}
		\matr{\sigma}^* = \frac{1}{\rho V_{ref}^2} \matr{\sigma} = -p^* \vekt{I} + \frac{1}{\reynolds} 
		\bigg(\nabxd\vel^* + \big(\nabxd\vel^*\big)^T\bigg) \,,
	\end{equation}
	\begin{subequations}
		\label{eq:nav-incomp-dim}
		\begin{equation}
			\nabxd \cdot \vel^* = 0 \,,
		\end{equation}
		\begin{equation}
			\label{eq:nav-incomp-dim-mom}
			\big(\vel^* \cdot \nabxd\big)\vel^* = \nabxd \cdot \matr{\sigma}^* \,,
		\end{equation}
	\end{subequations}
	where
	\begin{equation}
		\reynolds = \frac{\rho V_{ref} L_{ref}}{\mu} 
	\end{equation}
	is the dimensionless Reynolds number \cite{solid_book.ch1}.
	The Reynolds number describes the ratio between internal forces caused by inertia and viscous forces \cite{fef_book_ch1}. 
	
	\section{Physically Informed Neural Networks}\label{sec:pinn}
\begin{figure}[h!]
		\centering
		\includegraphics[width=\textwidth,scale=1.0]{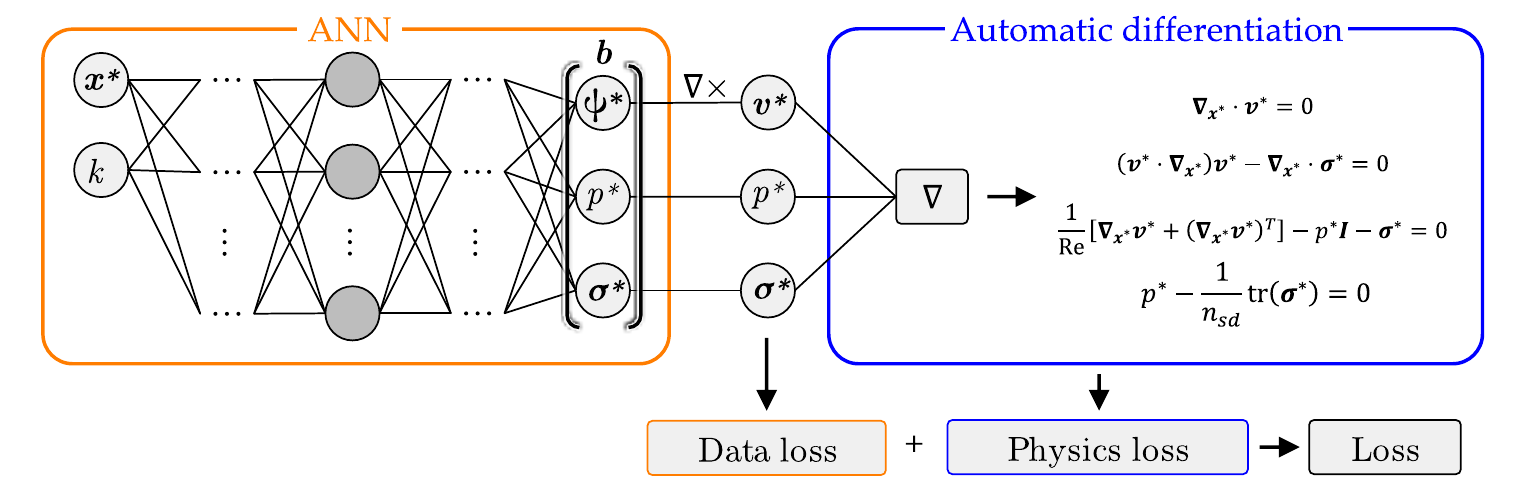}
		\caption[Schematic of PINN architecture]{Schematic of the PINN architecture.}
		\label{fig:PINN}
	\end{figure}

	Fundamentally, the PINN design is an approach to the training of a classical Artificial Neural Network (ANN).
	This is achieved through extending the ANN by a physics loss as depicted in \reffig{fig:PINN}.
	A feed-forward ANN is used as a surrogate model approximating the flow field in the form of the output vector $\vekt{b}(\xnew^*,...)$ as a function of space.
	The physics-informed network is constructed by adding the violation of the PDE as an additional
	loss term \cite{RAISSI2019686}.
	The necessary gradients w.r.t the input variables are computed by using automatic differentiation analogously to the backpropagation procedure \cite{RAISSI2019686,automaticdiff}.
	The physics-informed network thus shares the parameters with the ANN and its output is the physics residual $\vekt{\mathcal{R}}(\xnew^*,...)$.
	The physics loss $\mathcal{L}_f$ is then defined  as the corresponding mean squared error (MSE) loss and evaluated at a set of collocation points $\eta_f$.
	No labeled data has to be available at these points.
	Furthermore, a data loss is evaluated for two labeled datasets representing BC.
	A Neumann set $\eta_N$ is defined for which the pressure is prescribed.
	Analogously, the velocity error is evaluated for the Dirichlet set $\eta_D$.
	The total loss reads
	\begin{equation}
		\label{eq:loss}
		\mathcal{L} = f_{BC} \big( \mathcal{L}_D + \mathcal{L}_N \big) + \mathcal{L}_f \,.
	\end{equation}
	The weighting $f_{BC}$ is introduced to emphasize the adherence to the BC during training. Most previous works rely on stochastically sampling collocation points from a mathematically defined input domain \cite{RAISSI2019686,RAO2020207}.
	Nevertheless, this is impractical for geometrically complex scenarios.
	In the context of this work the required point sets are generated from finite-volume meshes using Siemens STAR-CCM+.
	Unlike conventional CFD simulations, PINN do not require good grid quality in terms of positioning of the collocation points.
	Nevertheless, it needs to be ensured that the number of collocation points available in a region are sufficient to resolve  expected local gradients.
	Rao et al. \cite{RAO2020207} propose a mixed-variable scheme using the Cauchy stress
	formulation of the incompressible Navier-Stokes equations.
	Improved learning performance is shown for two-dimensional flows by avoiding second derivatives of $\vel$ inside the loss formulation.
	The design presented as part of this work is therefore based on this mixed-variable scheme.\\

	The network input is given by the position $\xnew^*$ as well as an optional parameter $k$ for the parametric training of geometry variants.
	The stream function $\vekt{\Psi}^*$, the pressure and the Cauchy stress are chosen as the outputs of the ANN.
	The stream function is employed instead of the velocity to implicitly adhere to the mass conservation \cite{RAO2020207}.
	In three spatial dimensions a function that implicitly fulfills the divergence free condition can be defined as a three-component vector.
	The definition is shown in \refeqn{eq:streamfunc3d}.
	\begin{equation}
		\label{eq:streamfunc3d}
		\vekt{\nab{\xnew}} \times \vekt{\Psi} = \begin{bmatrix}
			\pderiv{\Psi_3}{y} - \pderiv{\Psi_2}{z} \vspace{0.1cm}\\
			\pderiv{\Psi_1}{z} - \pderiv{\Psi_3}{x} \vspace{0.1cm}\\
			\pderiv{\Psi_2}{x} - \pderiv{\Psi_1}{y} \\
		\end{bmatrix} = \vel  \,.
	\end{equation}
	The symmetric Cauchy stress tensor has six unique components.
	These are chosen as the network outputs.
	The complete output vector is thus given by \refeqn{eq:outputs3d}.
	\begin{equation}
		\label{eq:outputs3d}
		\vekt{b}(\xnew^*,k) = \begin{bmatrix}
			\Psi^*_1 &
			\Psi^*_2 &
			\Psi^*_3 &
			p^*_f &
			\sigma^*_{11} &
			\sigma^*_{12} &
			\sigma^*_{13} &
			\sigma^*_{23} &
			\sigma^*_{22} &
			\sigma^*_{33}
		\end{bmatrix}^T \,.
	\end{equation}
	The residual is constructed using the momentum conservation given by \refeqn{eq:cauchyDim} and \refeqn{eq:nav-incomp-dim-mom}.
	This yields \refeqn{eq:residualPINN}.
	\refeqn{eq:residualtrace} holds for the given divergence free velocity field.
	It is introduced as an additional relation between the stress components and pressure.
	The physics loss therefore reads as in \refeqn{eq:physicsloss}, where $\mathcal{L}_{\bs{v}}$, $\mathcal{L}_{\bs{\sigma}}$ and $\mathcal{L}_{p}$ are the MSE loss of the respective residual components.
	The stress loss weighting factor denoted by $f_\sigma$ is introduced to improve the training with regard to shear stresses.
	The related loss contribution becomes small for large Reynolds numbers which may lead
	to a high relative violation and thus inaccurate approximation.
	\begin{subequations}
		\label{eq:residualPINN}
		\begin{empheq}[left={\vekt{\mathcal{R}}(\vel^*,p^*,\matr{\sigma}^*) = \empheqlbrack},
			right={\empheqrbrack \stackrel{!}{=} \vekt{0} \,.}]{gather}
			\big(\vel^* \cdot \nabxd\big)\vel^* - \nabxd \cdot \matr{\sigma}^* \label{eq:residualmom}\\
			\frac{1}{\reynolds} \big(\nabxd\vel^* + \big(\nabxd\vel^*\big)^T\big) - p^* \vekt{I} -
			\matr{\sigma}^* \label{eq:residualstress}\\
			p^* + \frac{1}{n_{sd}} \textrm{tr}(\matr{\sigma}^*) \label{eq:residualtrace}
		\end{empheq}
	\end{subequations}
	\begin{equation}
		\label{eq:physicsloss}
		\mathcal{L}_{f} = \mathcal{L}_{\bs{v}}(\mathcal{R}_1) + f_\sigma
		\mathcal{L}_{\bs{\sigma}}(\mathcal{R}_2) +
		\mathcal{L}_{p}(\mathcal{R}_3)  \,.
	\end{equation}
	To quantify the approximation accuracy, a test loss $\mathcal{L}_{test}$ is defined.
	In every presented case, 1\,\% of all collocation points are excluded from the model training to create a test set.
	The error is evaluated relative to reference solutions obtained through CFD simulations using an identical grid.
	The relative L2 error in $\vel$ and $p$ is used and reads
	\begin{equation}
		\mathcal{L}_{test} = \frac{\sqrt{\sum\limits_{\eta_{test}} \sum\limits_{\hat{x}\in\{v_i,p\}} \vert
				\hat{x}_{pred} - \hat{x}_{ref}\vert^2 }}{\sqrt{\sum\limits_{\eta_{test}}
				\sum\limits_{\hat{x}\in\{v_i,p\}} \vert \hat{x}_{ref}\vert^2}}  \,,
	\end{equation}
	where $\eta_{Test}$ is the test set.
	This exclusion has no significant impact on the density of training points but provides an estimate of the prediction accuracy within the considered domain.
	
	\section{Parametric Training Algorithm}
	\label{sec:trainingAlg}
	
	Training is performed in two segments. As proposed by Rao et al. a preconditioning is performed using the Adam optimizer \cite{RAO2020207,adam}. 
	Subsequently the model is trained utilizing the L-BFGS optimizer \cite{lbfgs} to exploit its fast convergence.
	The implemented training algorithm operates on multiple randomized subsamples of the data (minibatches) by dividing the datasets according to a maximum
	batch size.
	The introduction of an input parameter aside from the spatial position yields the challenge of efficiently generating a collocation point set spanning the entire input domain.
	Existing works introduce time as an additional input dimension by stochastically sampling a larger set of collocation points in all dimensions \cite{RAISSI2019686,RAO2020207,HAGHIGHAT2021113552,MAO2020112789}.
	This, however, increases the dataset size significantly whilst, for largely static boundary conditions, carrying little additional information.
	The aim of this work is the introduction of geometric variation.
	This demands a spatial change of the generated point set depending on the parameter value which is non-trivial.\\
	
\begin{figure}[t]
		\centering
		\includegraphics[width=\textwidth]{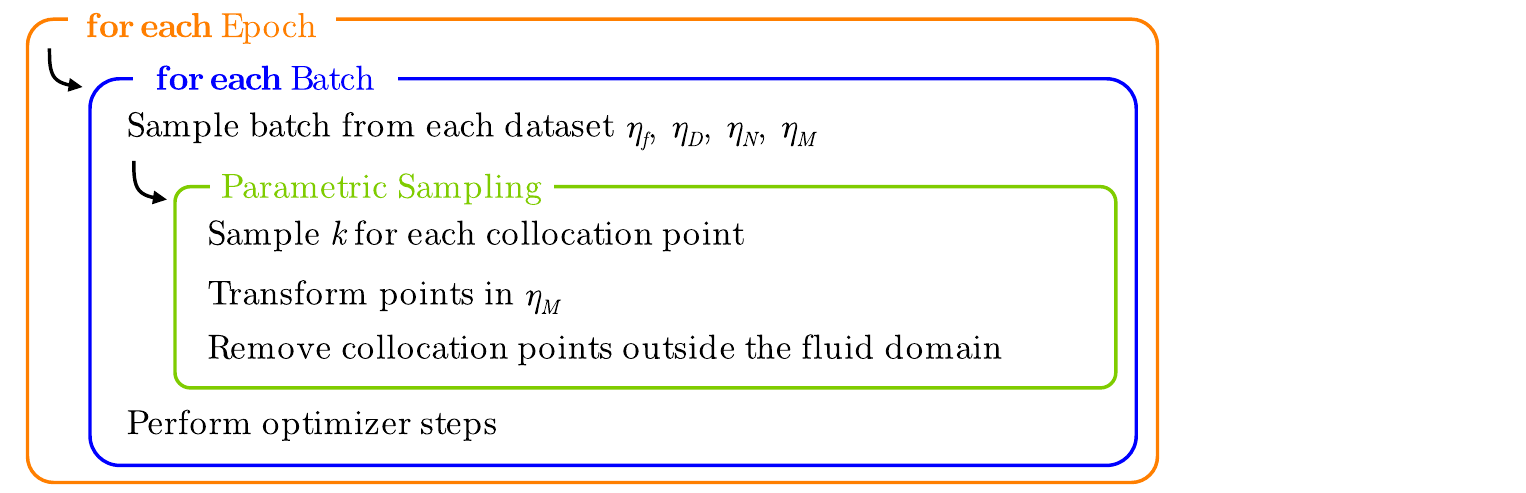}
		\caption[Parametric training algorithm]{Parametric training algorithm.}
		\label{fig:paramAlg}
	\end{figure}

	The newly proposed algorithm, depicted in \reffig{fig:paramAlg}, aims to work on a single set of points generated from a spatial mesh.
	The points are still distributed into the previously defined groups of volume, Dirichlet boundary and Neumann boundary points.
	Furthermore, an additional point set $\eta_M$ is introduced to distinguish between Dirichlet boundary points whose position depends on the respective parameter setting and those that are non-moving Dirichlet boundary points.
	Thus, all points within $\eta_f$, $\eta_D$ and $\eta_N$ are static
	whereas points within $\eta_M$ represent wall translations according to the assigned parameter.
	In Equations \eqref{eq:paramAlg} the algorithm is expressed as a set of operations applied to each point set.\\
	
\begin{figure}[h!]
		\centering
		\begin{tabular}{ll}
			A: & \\
			\includegraphics[scale=0.8]{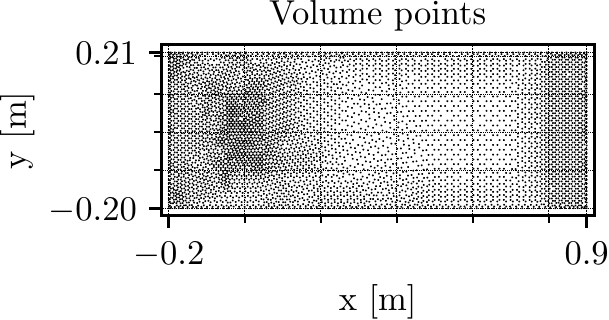} &
			\includegraphics[scale=0.8]{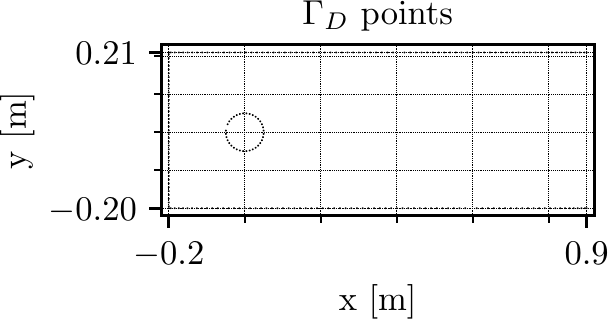}\\
			B: & \\
			\includegraphics[scale=0.8]{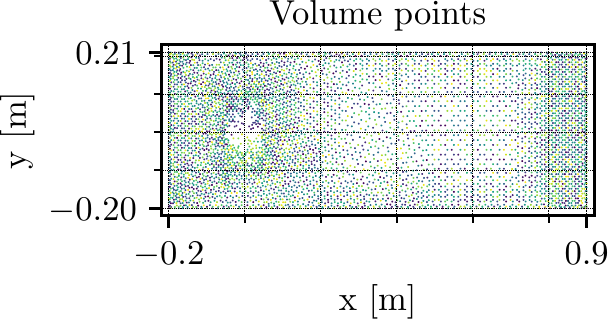} &
			\includegraphics[scale=0.8]{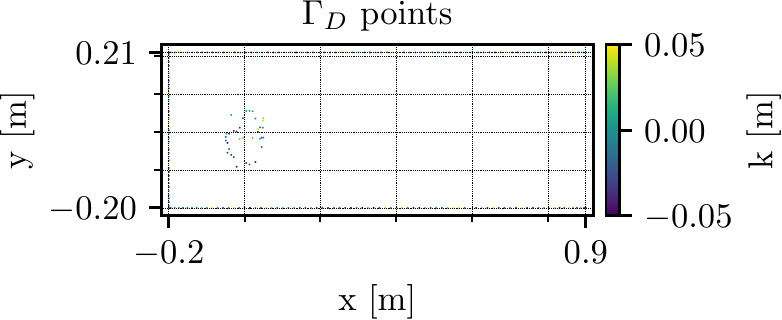}
		\end{tabular}
		\caption[Parameter sampling during training]{Exemplary representation of the parameter
			sampling during training of the two-dimensional cylinder scenario. A: Imported mesh, B:
			transformed mesh after sampling.}
		\label{fig:sampling}
	\end{figure}

	Initially the 
	$\textrm{\textit{sample}}:\xnew\mapsto\begin{bmatrix}
		\xnew & k
	\end{bmatrix}^T$
	operation is introduced to assign values for $k$.
	Parameter values are randomly sampled from the allowed parameter space for each collocation point individually.
	As a result the PINN is  trained for a continuous parameter space instead of using an additional discrete mesh for each configuration.
	Furthermore, the proposed PINN class implements three methods specific to the respective
	scenario.
	$\textrm{\textit{transform}}(\xnew,k)$ translates the points in $\eta_M$ according to the corresponding value of $k$. $\textrm{\textit{inside}}_{fDN}(\xnew,k)$ and $\textrm{\textit{inside}}_M(\xnew,k)$ represent logical conditions that are true if $\xnew$ is inside the spatial domain for the given parameter value and false otherwise.
	Points outside the domain are excluded from training. 
	This results the transformed point set $\overline{\overline{\eta}}$.
	\begin{subequations}
		\label{eq:paramAlg}
		\begin{align}
			\overline{\eta}_i &= \textrm{\textit{sample}}(\eta_i) \,, & i\in\{f,D,N,M\} \,,\\
			\tilde{\eta}_M &= \textrm{\textit{transform}}(\overline{\eta}_M)  \,,& \\
			\overline{\overline{\eta}}_i &= \textrm{\textit{inside}}_{fDN}(\overline{\eta}_i) \,, & i\in\{f,N\} \,,
			\label{eq:paramAlgInsidefN}\\
			\overline{\overline{\eta}}_D &= \textrm{\textit{inside}}_{fDN}(\overline{\eta}_D) +
			\textrm{\textit{inside}}_M(\tilde{\eta}_M)  \,.& \label{eq:paramAlgInsideDM}
		\end{align}
	\end{subequations}
	An illustration of the parameter sampling during training is given by \reffig{fig:sampling} for the
	cylinder scenario presented in \refsec{sec:cyl3dp}.
	The static datasets are required to include collocation points spanning the entire spatial domain relevant for the respective parameter space.
	The presented scenario uses a maximum cylinder displacement of $k \pm 0.05\munit{\,m}$.
	Therefore, only volume points are removed during sampling.
	The Dirichlet boundary is formed by the sets $\eta_D$ and $\eta_M$ whereas the latter is transformed to the cylinder's vertical position.
	
	\section{Flow Prediction via PINN for 3D Domains}\label{sec:3dresults}
	The extension of the proposed approach to the forward prediction of flow solutions from two into three 
	spatial dimensions poses a computational challenge. Although readily available from a 
	mathematical perspective, the governing equations rise in complexity. The number of trained 
	network outputs is doubled by the addition of the third dimension. Furthermore, the number of 
	necessary collocation points to achieve a comparable resolution of three instead of two spatial 
	dimensions is increased significantly. The nondimensional formulation has therefore been developed using non-parametric (static), two-dimensional problems before introducing parametric training and a third spatial dimension.
	In the following, the performance of the three-dimensional training is first evaluated on a non-parametric 
	problem in \refsec{sec:cyl3d}. Subsequently, the parametric training results are presented for two 
	geometric scenarios.
	The test strategy for all problems involves a grid search to determine the optimal network configuration as well as loss weighting factors $f_{BC}$ and $f_{\sigma}$. The best results and relevant effects regarding the configuration or convergence behavior are then presented.

	\subsection{Static Cylinder: Minibatch Algorithm and Multivariate Scheme}
	\label{sec:cyl3d}
	
\begin{figure}[h!]
		\centering
		\begin{minipage}[c]{8.0cm}
			\includegraphics[width=\textwidth,scale=1.0]{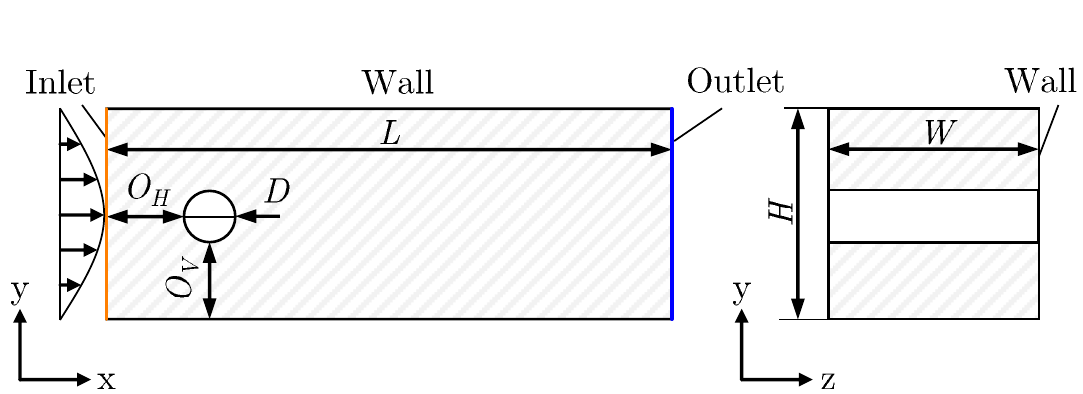}
		\end{minipage}
		\begin{minipage}[c]{3.4cm}
			\begin{tabular}{c|r}
				\toprule
				Name & Length $[\munit{m}]$ \\
				\midrule
				$L$ & $1.1$\\
				$H$ & $0.41$\\
				$O_H$ & $0.15$\\
				$O_V$ & $0.1$ - $0.2$\\
				$D$ & $0.1$\\
				$W$ & $0.4$\\
				\bottomrule
			\end{tabular}
		\end{minipage}
		\caption[Configuration of the two-dimensional cylinder scenario]{Configuration of the 
			two-dimensional cylinder scenario.}
		
		\label{fig:mapCyl2d}
	\end{figure}

	The first scenario is based on the investigations of Rao et al. \cite{RAO2020207}.
	A cylinder is placed inside a parallel flow as depicted in \reffig{fig:mapCyl2d}. A bi-quadratic inflow 
	velocity profile with a maximum velocity of $v_{in,max} = 1.0\munit{\,\frac{m}{s}}$ is defined as: 
	\begin{equation}
		\label{eq:vel_profile}
		v_{in}(y,z) = \frac{16}{H^2 W^2}\big(\frac{H}{2} - y\big) \big(\frac{H}{2} + y\big) \big(W - z\big) z \,.
	\end{equation}
	A pressure outlet is defined while a no-slip condition is imposed at the cylinder and tunnel walls. The fluid
	density and viscosity are set to $\rho=1.0\munit{\,\frac{kg}{m^3}}$ and $\mu = \expnum{2}{-2}\munit{\Pas}$ respectively. 
	
\begin{figure}[h!]
		\centering
		\includegraphics[width=\textwidth,scale=1.0]{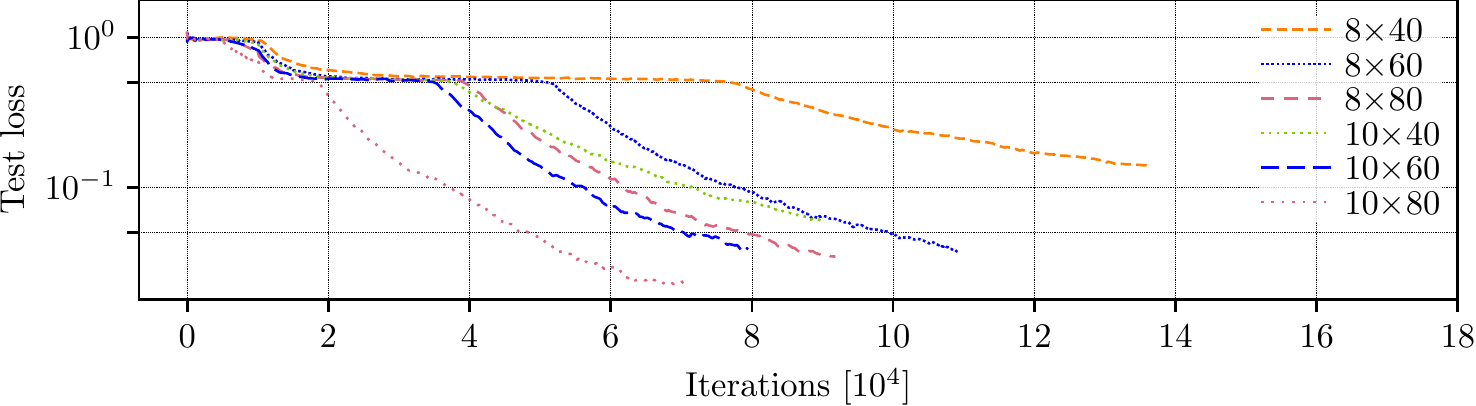}
		\caption[Convergence curves for a three-dimensional, static cylinder for different network 
		configurations]{Convergence curves for a three-dimensional, static cylinder for different network 
			configurations. Total iterations count Adam iterations as well as L-BFGS inner iterations for each 
			batch.}
		\label{fig:cyl3dtestlosses}
	\end{figure}

	A collocation point set of $N_f = 
	57813$ volume points, $N_D = 14716$ Dirichlet points and $N_N = 1725$ Neumann boundary 
	points is utilized to train the PINN without geometric variation. 
	The dataset is split into two batches. BC and stress loss weighting factors of $f_{BC} = 10$ and 
	$f_\sigma = 1$ are applied respectively. According to the chosen BC reference parameters of $L_{ref} = 
	1.1\munit{\,m}$ and $V_{ref} = 1.4\munit{\,\frac{m}{s}}$ are used to scale the labeled data.
	The optimal network configuration is  determined by a grid search strategy.
	As a result of the grid search, the final test loss as well as the average elapsed time per optimizer step (SIET), shown in table \reftab{tab:cyl3dsettings}, are compared for the different network configurations.
	The training of each model terminates whenever the L-BFGS optimizer cannot determine a descent direction anymore and has thus reached a local optimum.
	The extension of the network width from $n=40$ to $n=60$ in particular reduces the relative error 
	below $\mathcal{L}_{test} = 0.05$. 
	The SIET generally increases with the network capacity while the number of necessary iterations 
	decreases as shown in \reffig{fig:cyl3dtestlosses}. 
	The optimal total training time is thereby obtained for the configurations  $n\times m = 10\times 
	40/60$. The configuration using $n=10$ layers with $m=60$ neurons is chosen for further 
	investigations as is shows the best overall performance.
	The training iteration always  converges to a stable minimum, suggesting that the gradient 
	constraints imposed by the PDE pose a sufficient regularization mechanism.
	
	\begin{table}[h!]
		\centering
		%\begin{minipage}{.5\linewidth}
		\caption[Static cylinder test loss for different PINN configurations]{Test loss and SIET for 
			different PINN configurations and the training of a static, three-dimensional cylinder. \textit{Left (right):} 
			test loss (SIET).}
		\begin{tabular}{cccc}
			\toprule
			$n\ \backslash\ m$ & $40$ & $60$ & $80$ \\
			\midrule
			$8$ & 0.137 (0.155$\munit{\,s}$) & 0.037 (0.198$\munit{\,s}$) & 0.034 (0.275$\munit{\,s}$) \\
			$10$ & 0.059 (0.189$\munit{\,s}$) & 0.037 (0.250$\munit{\,s}$) & 0.022 
			(0.337$\munit{\,s}$) \\
			\bottomrule
		\end{tabular}
		\label{tab:cyl3dsettings}
		%\end{minipage}
	\end{table}

	%\begin{remark}
		\paragraph*{Remark:} \textit{To provide comparability, in all plots where PINN results are compared with CFD simulations, the CFD solution is interpolated to the transformed set of collocation points. Prior to this, the CFD domain is adjusted to match the change in geometry included in the PINN.}
	%\end{remark}

\begin{figure}[h!]
		\centering
		\begin{tabular}{lll}
			\toprule
			CFD & PINN & Difference \\
			\midrule
			\includegraphics[scale=1.0]{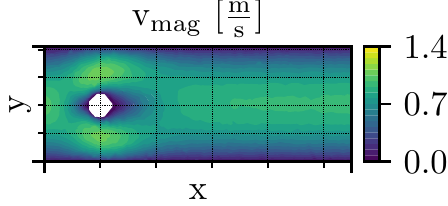} &
			\includegraphics[scale=1.0]{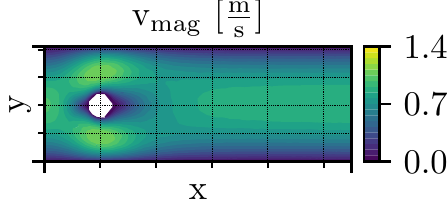} & 
			\includegraphics[scale=1.0]{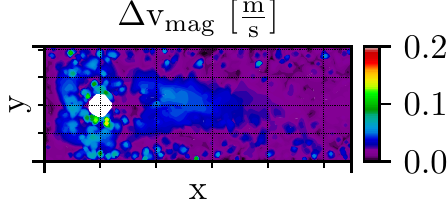} \\
			\includegraphics[scale=1.0]{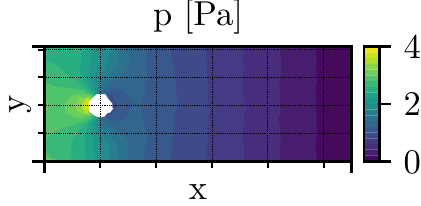} & 
			\includegraphics[scale=1.0]{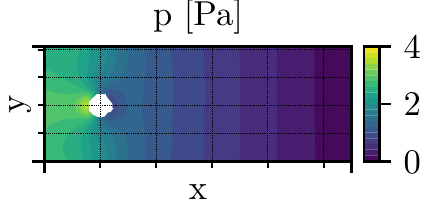} & 
			\includegraphics[scale=1.0]{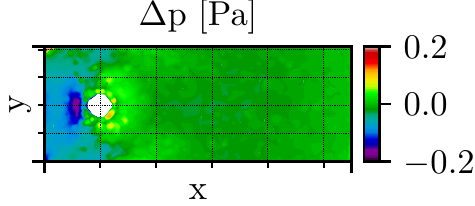}\\
			\includegraphics[scale=1.0]{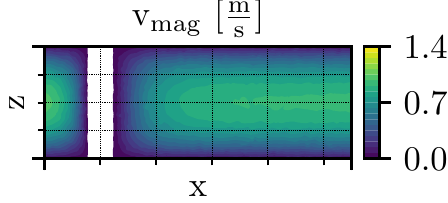} & 
			\includegraphics[scale=1.0]{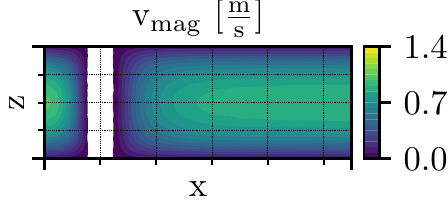} & 
			\includegraphics[scale=1.0]{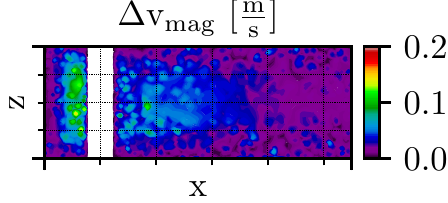} \\
			\includegraphics[scale=1.0]{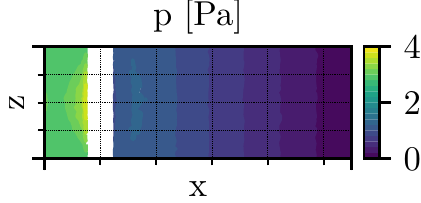} & 
			\includegraphics[scale=1.0]{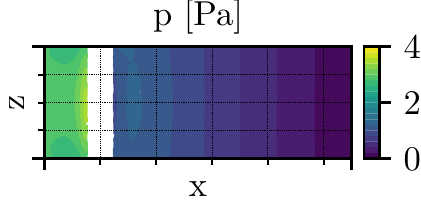} & 
			\includegraphics[scale=1.0]{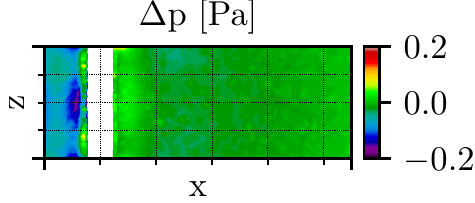}\\
			\bottomrule
		\end{tabular}
		\caption{xy- and xz-cuts at $z=0.2\munit{\,m}$ and $y=0\munit{\,m}$ of a three-dimensional flow around a cylinder. ${\Delta v}_{mag}$ denotes the magnitude of the velocity error. Used model settings: $n=10,~m=60,~f_{BC}=10$ and $f_{\sigma}=1$.}
		\label{fig:cyl3d_E2}
	\end{figure}

	In \reffig{fig:cyl3d_E2} the obtained PINN prediction is compared to a CFD reference solution. The 
	velocity and the pressure field are shown to be in good agreement with the reference. The effects 
	of the added $z$-dimension are captured well. The maximum velocity error occurs close to the 
	cylinder. It does not exceed ${\Delta v}_{mag} = 0.15\munit{\,\frac{m}{s}}$. The high-pressure field 
	around the front stagnation point is predicted well, including a pressure drop towards the tunnel 
	side wall. A maximum pressure of $p=3.7\munit{\,Pa}$ occurs at the front center of the cylinder. At 
	this point the maximum pressure error is observed, corresponding to an underprediction of $\Delta 
	p=-0.2\munit{\,Pa}$.
	
\begin{figure}[h!]
		\centering
		\begin{tabular}{lll}
			\toprule
			No $\vekt{\Psi}$ & No $\matr{\sigma}$ & Mixed, full batch  \\
			\midrule
			\includegraphics[scale=1.0]{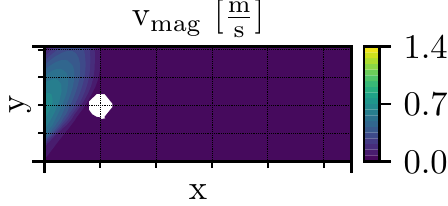} & 
			\includegraphics[scale=1.0]{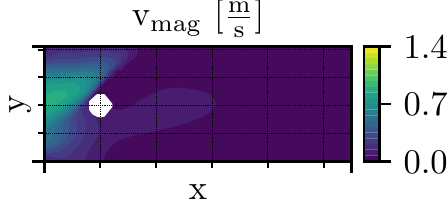} & 
			\includegraphics[scale=1.0]{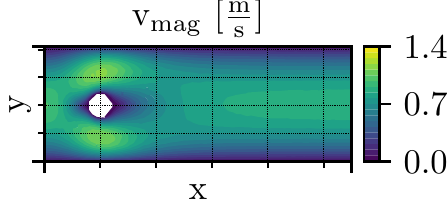} \\
			\includegraphics[scale=1.0]{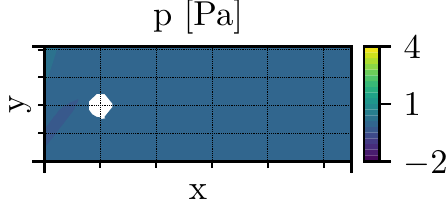} & 
			\includegraphics[scale=1.0]{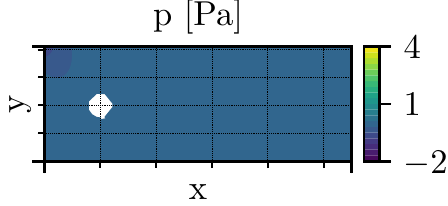} & 
			\includegraphics[scale=1.0]{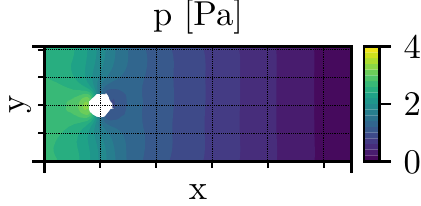}\\
			\bottomrule
		\end{tabular}
		\caption{xz-cuts of a three-dimensional flow around a cylinder with model settings: $n=10,~m=60,~f_{BC}=10$ and $f_{\sigma}=1$. Comparison of obtained pressure and velocity fields for different training schemes. \textit{Left:} No stream function as output, \textit{Center:} No Cauchy stress as output, \textit{Right:} Full batch training of mixed variable scheme.}
		\label{fig:cyl3d_E2_formulations}
	\end{figure}

\begin{figure}[h!]
		\centering
		\includegraphics[width=\textwidth,scale=1.0]{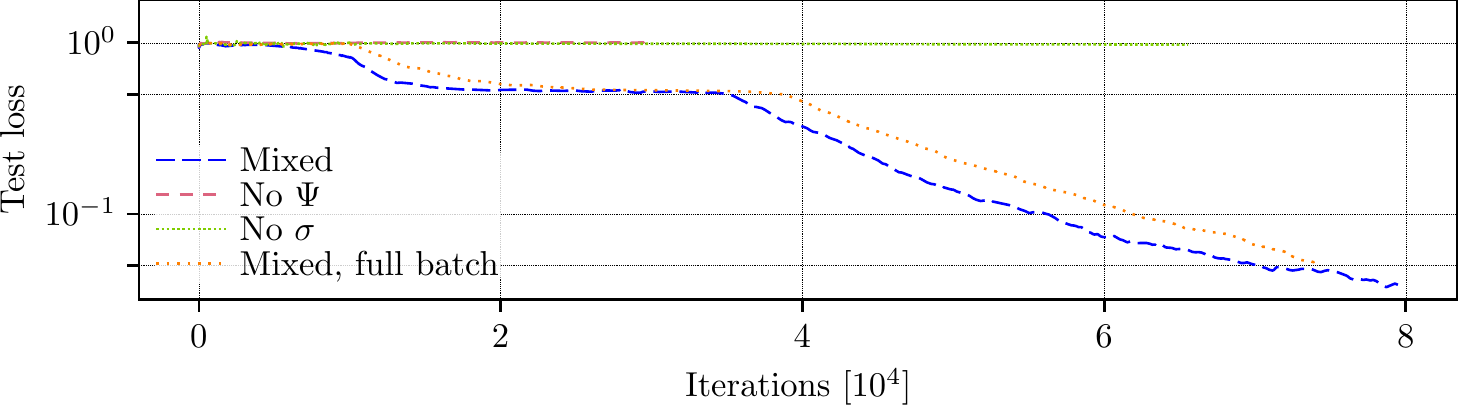}
		\caption[Convergence curves for a three-dimensional, static cylinder and different loss 
		formulations]{Convergence curves for a three-dimensional, static cylinder and different loss 
			formulations at model settings: $n=10,~m=60,~f_{BC}=10$ and $f_{\sigma}=1$. Total iterations count Adam iterations as well as L-BFGS inner iterations for each 
			batch.}
		\label{fig:cyl3dformulationsLosses}
	\end{figure}

	The PINN is also trained using different formulations of the physics loss as well as full batch 
	training. The obtained results are depicted in \reffig{fig:cyl3d_E2_formulations} with the 
	corresponding convergence curves being plotted in \reffig{fig:cyl3dformulationsLosses}.
	Rao et al. show that a PINN is unable to recover the flow field around a two-dimensional cylinder if 
	the velocity is used to compute the viscous terms within the momentum equation instead of a 
	separate stress output \cite{RAO2020207}. These results are confirmed for the three-dimensional 
	case. The left and center column of \reffig{fig:cyl3d_E2_formulations} show the learned flow fields 
	for training without the use of the stream function or the Cauchy stress formulation respectively. 
	For both formulations no physically plausible flow field is learned and the test loss depicted in 
	\reffig{fig:cyl3dformulationsLosses} is not reduced significantly. The imposed BC are not captured 
	well with the bulk of the inflow leaving the domain through the top wall in front of the cylinder. No 
	pressure gradient is present and the entire pressure field thus exhibits the zero value prescribed at 
	the outlet boundary.
	
	The convergence curve for the proposed mixed variable scheme and training in full batch 
	configuration is also shown. It is comparable to the minibatch training performed with two batches. 
	The minibatch training shows a measurable advantage during the first $10000$ iterations of Adam 
	preconditioning. The error is reduced visibly while no reduction occurs in full batch training. Aside 
	from the difference carried over from the preconditioning, both configurations perform similarly 
	during L-BFGS training. Convergence is reached within $80000$ total iterations. As these count 
	inner iterations for each batch separately the number of performed epochs is in fact halved for the 
	minibatch training. Therefore, the loss reduction per pass over the whole dataset is doubled. The 
	full batch training converges to a test loss of 
	$\mathcal{L}_{test}\approx\expnum{5.2}{-2}$. Using the minibatch algorithm this value is reached 
	about 3000 iterations earlier. The error then plateaus for about 3000 iterations before decreasing 
	to a minimum of $\mathcal{L}_{test}\approx\expnum{3.7}{-2}$. As both configurations converge 
	well, the predicted flow fields are in good agreement as shown in \reffig{fig:cyl3d_E2} and 
	\reffig{fig:cyl3d_E2_formulations}$\,$(right).
	For the 2 batch minibatch procedure the SIET is increased by about $7\,\%$ relative to full batch 
	training. Due to the reduced number of necessary epochs the total training time is reduced by 
	$38.1\,\%$. The maximum GPU memory usage occurs during the L-BFGS training procedure. It is reduced by $37.8\,\%$ compared to the single batch requirement. 
	
	\subsection{Parametric Cylinder in Parallel Flow}
	\label{sec:cyl3dp}
	
	This section studies the possibility to extend the cylinder scenario introduced in the previous 
	section by a geometric parameter. A vertical cylinder displacement $k\in [-0.05,0.05]\munit{\,m}$ 
	is added as the input parameter. A collocation point set with $N_f = 66622$ total volume points is 
	utilized. 
	The points are refined around the possible cylinder positions. The static boundary is resolved by 
	$N_D = 14716$ Dirichlet points and $N_N = 1725$ Neumann boundary points.
	The operations $\textrm{\textit{inside}}_{fDN}(\xa,k)$, $\textrm{\textit{inside}}_M(\xa,k)$ and 
	$\textrm{\textit{transform}}(\xa,k)$ are defined as:
	\begin{subequations}
		\label{eq:cyl3dPfunctions}
		\begin{align}
			\textrm{\textit{inside}}_{fDN} &: \xa,k\mapsto \vert\vert \begin{bmatrix}
				x \\ y - k \\ 0
			\end{bmatrix} \vert\vert_2 \geq \frac{1}{2}D \,,\\
			\textrm{\textit{inside}}_{M} &: \xa,k\mapsto \textrm{True} \,,\\
			\textrm{\textit{transform}} &: \xa,k\mapsto \xa + \begin{bmatrix}
				0 \\ k \\ 0
			\end{bmatrix} \,.
		\end{align}
	\end{subequations}

\begin{figure}[h!]
		\centering
		\begin{tabular}{lll}
			\toprule
			$k=-0.05\munit{\,m}$ & $k=0.0\munit{\,m}$ & $k=0.03\munit{\,m}$ \\
			\midrule
			\includegraphics[scale=1.0]{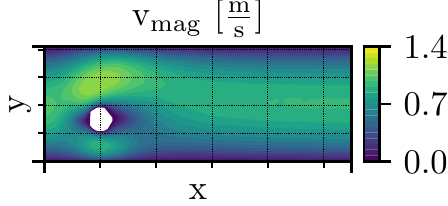} & 
			\includegraphics[scale=1.0]{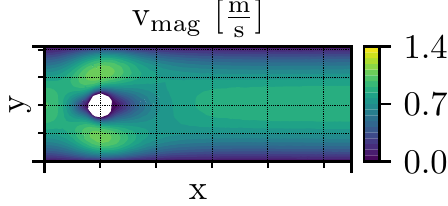} & 
			\includegraphics[scale=1.0]{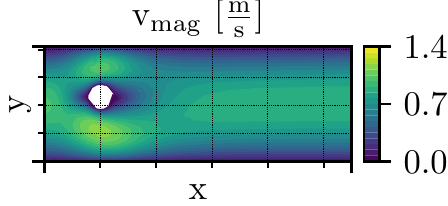} \\
			\includegraphics[scale=1.0]{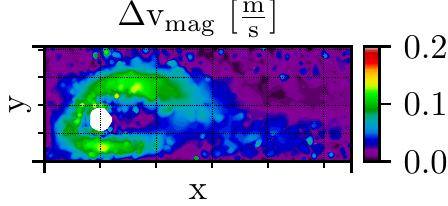} & 
			\includegraphics[scale=1.0]{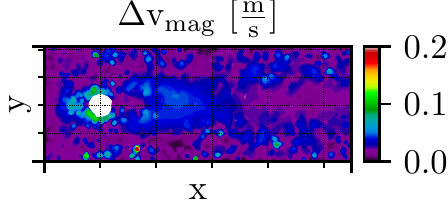} & 
			\includegraphics[scale=1.0]{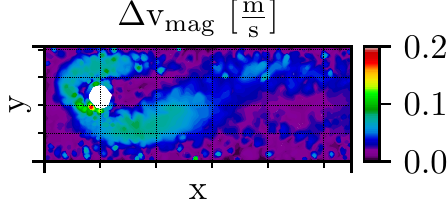} \\
			\includegraphics[scale=1.0]{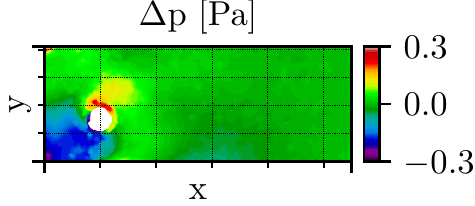} & 
			\includegraphics[scale=1.0]{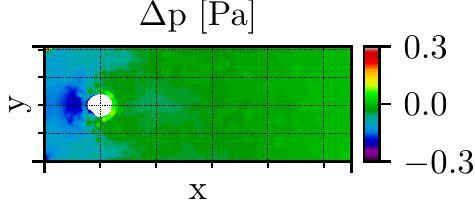} & 
			\includegraphics[scale=1.0]{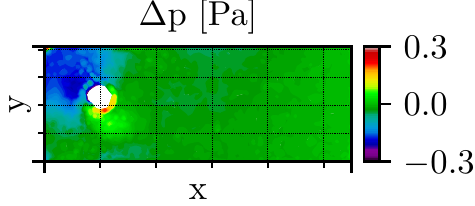} \\
			\bottomrule
		\end{tabular}
		\caption{xy-cuts at $z=0.2\munit{\,m}$. Velocity and pressure error of a 
			two-dimensional, parametric flow passing a cylinder, using the network configuration $n\times m = 10\times 60$.}
		\label{fig:cyl3dP_E2_diffs}
	\end{figure}

	All BC and the loss weighting factors are kept equal to the static configuration.
	The obtained results are illustrated in \reffig{fig:cyl3dP_E2_diffs} on the example of an xy-cut 
	through the cylinder center. 
	The first row shows the velocity field for the three cylinder displacements $k\in\{-0.05,0.0,0.03\}\munit{\,m}$. The flow shifts from predominantly passing above the cylinder to passing below for increasing $k$. 
	The second and third row depict the corresponding velocity and pressure error evaluated at the grid nodes. The predictions for each $k$ are compared to three corresponding CFD reference computations. The velocity error is similar to the static training close to the 
	center of the parameter space but increases towards the edges. 
	For $k\neq0\munit{\,m}$ the PINN underpredicts the change in velocity next to the cylinder. While 
	the velocity error does not surpass $0.05\munit{\,\frac{m}{s}}$ for $k=0\munit{\,m}$ the maximum 
	error increases to $\Delta v_{max}\approx 0.12\munit{\,\frac{m}{s}}$ for $k=0.05\munit{\,m}$. This 
	corresponds to a maximum relative error of $9.5\,\%$ compared to the maximum flow velocity next 
	to the cylinder.
	
	\subsection{Parametric T-Junction}
	\label{sec:tjunc3dp}
	
\begin{figure}[t]
		\centering
		\begin{minipage}[c]{8cm}
			\includegraphics[width=\textwidth,scale=1.0]{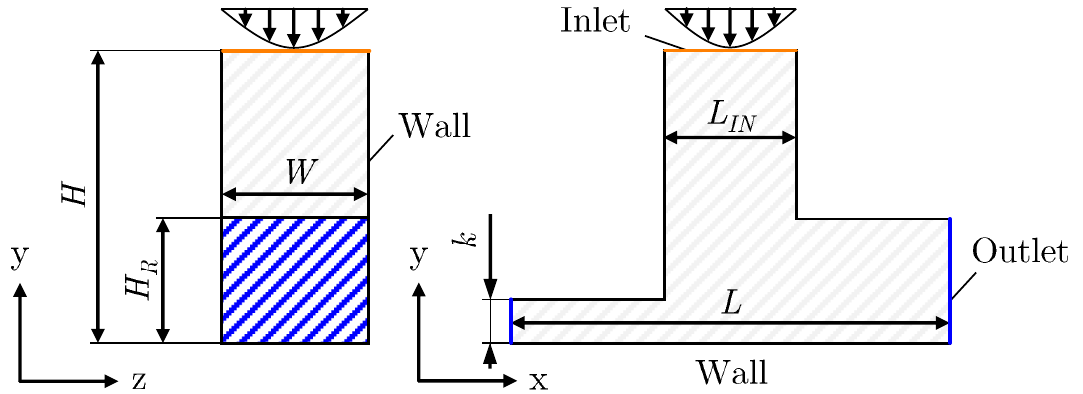}
		\end{minipage}
		\begin{minipage}[c]{3.4cm}
			\begin{tabular}{c|r}
				\toprule
				Name & Length $[\munit{m}]$ \\
				\midrule
				$L$ & $0.3$\\
				$H$ & $0.2$\\
				$L_{IN}$ & $0.09$\\
				$k$ & $0.03$ - $0.07$\\
				$H_R$ & $0.08$\\
				$W$ & $0.1$ \\
				\bottomrule
			\end{tabular}
		\end{minipage}
		\caption[Configuration of the three-dimensional T-junction scenario]{Configuration of the 
			three-dimensional T-junction scenario.}
		\label{fig:mapTjunc3d}
	\end{figure}

\begin{figure}[H]
		\centering
		\includegraphics[width=\textwidth]{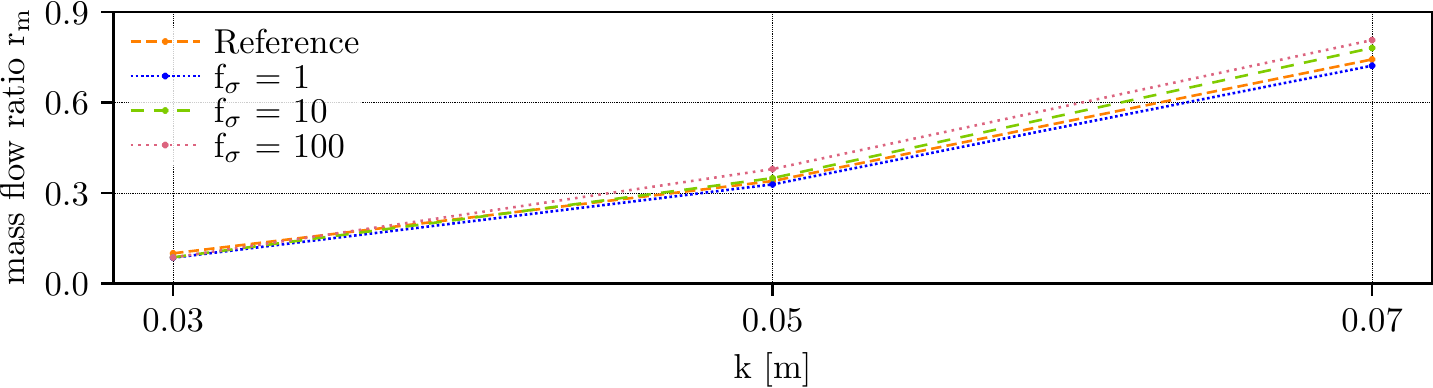}
		\caption[Mass flow ratio at the outlets of a three-dimensional T-junction]{Mass flow ratio at the 
			outlets of a three-dimensional T-junction plotted over different junction configurations.}
		\label{fig:tjunc3dp_ratio}
	\end{figure}

\begin{figure}[H]
		\centering
		\begin{tabular}{lll}
			\toprule
			CFD & PINN &   \\
			\midrule
			$k=0.03\munit{m}$ & & \\
			\includegraphics[scale=1.0]{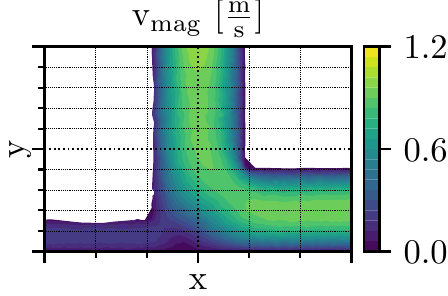} & 
			\includegraphics[scale=1.0]{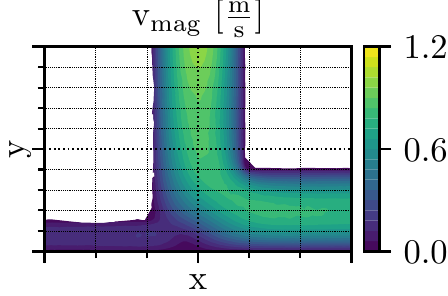} & 
			\includegraphics[scale=1.0]{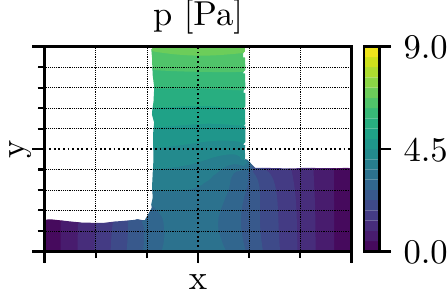} \\
			$k=0.07\munit{m}$ & & \\
			\includegraphics[scale=1.0]{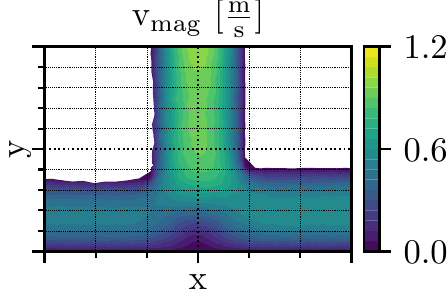} & 
			\includegraphics[scale=1.0]{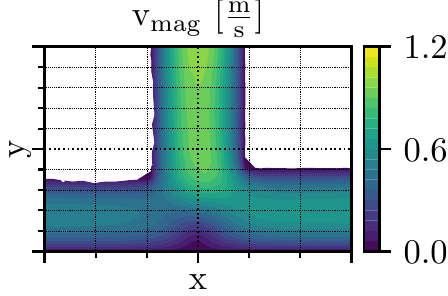} & 
			\includegraphics[scale=1.0]{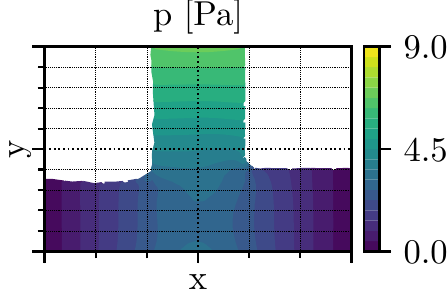}\\
			\bottomrule
		\end{tabular}
		\caption{xy-cuts of a three-dimensional flow inside a parametric T-junction compared to a CFD reference using the network configuration $n\times m = 10\times 60$, $f_{BC}=10$ and  $f_\sigma=1$.}
		\label{fig:tjunc3d_E2}
	\end{figure}

	The goal of the T-junction scenario is to test the PINN's ability to predict flows with significant 
	changes to the main flow direction by learning the parametric BC. \reffig{fig:mapTjunc3d} 
	illustrates the utilized junction setup.
	Analogously to the cylinder scenario and \refeqn{eq:vel_profile} a bi-quadratic velocity inlet profile with $v_{in,max} = 1.0 \munit{\,\frac{m}{s}}$ is prescribed at the inlet boundary with $\rho=1.0\munit{\,\frac{kg}{m^3}}$ 
	and $\mu = \expnum{2}{-2} \munit{\Pas}$. 
	A pressure outlet is defined at both junction arms while a no-slip condition is used at the junction 
	walls. The unequal height of the junction arms leads to an asymmetric flow field. For applications 
	such as the transport of heat it is of particular interest to predict the mass flow ratio if different 
	flow paths are available. Thus, the flow field as well as the predicted outflow ratios are investigated 
	for different height ratios. The left junction height is introduced as geometric variation $k\in 
	[0.03,0.07]\munit{\,m}$.
	The junction is resolved by a total of $N_f = 75487$ volume collocation points.
	BC are enforced at $N_N = 1807$ static Neumann boundary points and $N_D = 19725$ Dirichlet points.
	$N_M=3337$ of the Dirichlet points resolve the moving left junction. 
	Due to the reduced domain size the reference parameters for de-dimensionalization are chosen as $L_{ref} = 0.3\munit{\,m}$ and $V_{ref} = 1.0\munit{\,\frac{m}{s}}$.
	Training is performed using two batches.
	The definitions of the scenario specific operations are given by Equations
	\refeq{eq:tjunc2dPfunctions}. 
	\begin{subequations}
		\label{eq:tjunc2dPfunctions}
		\begin{align}
			\textrm{\textit{inside}}_{fDN} &: \xa,k\mapsto y \leq k \lor x \geq -\frac{1}{2}L_{IN}  \,,\\
			\textrm{\textit{inside}}_{M} &: \xa,k\mapsto y \leq H  \,,\\
			\textrm{\textit{transform}} &: \xa,k\mapsto \xa + \begin{bmatrix}
				0 \\ k - 0.03\munit{\,m}
			\end{bmatrix} \,.
		\end{align}
	\end{subequations}
	Resulting from another grid search, the PINN configuration of $n\times m=10\times60$ is utilized for the paramteric training.
	The predicted flows for $k\in\{0.03,0.07\}\munit{\,m}$ are illustrated in \reffig{fig:tjunc3d_E2} compared to CFD reference solutions.
	For $k=0.03\munit{\,m}$ the flow predominantly passes through the right junction whereas for  $k=0.07\munit{\,m}$ the flow is about symmetric.
	In both configurations the predicted flow is in good agreement with the CFD simulation.
	\reffig{fig:tjunc3dp_ratio} shows the observed outlet mass flow ratio $r_m = \frac{\dot{m}_L}{\dot{m}_R}$ for different stress weighting factors compared to the CFD reference.
	The approximation quality is tested using the factors $f_\sigma\in\{ 1,10,100\}$ whereas the BC loss factor is kept at $f_{BC} = 100$. 
	It can be seen that the effect of the geometric change is captured by the PINN solution.
	The mass flow ratio is roughly proportional to $k$ and shifts between a minimum of $10\,\%$ and a maximum of $80\,\%$. 
	The predicted ratio is in good agreement with the reference solutions irrespective of the stress loss factor. The error never exceeds $6\,\%$. 
	The maximum error occurs for a fully opened junction arm of $k=0.07\munit{\,m}$.
	Notably, while $r_m$ is underpredicted by a maximum of $2\,\%$ without stress loss scaling, the dependency is overpredicted if scaling is applied. 
	
	\section{PINN Applied in Thermal Coupling Design Process of an Exhaust System Scenario}\label{sec:exhaust}
\begin{figure}[h!]
		\centering
		\setlength{\fboxsep}{0pt}
		\setlength{\fboxrule}{0.5pt}
		\fbox{\includegraphics[height=3.0cm]{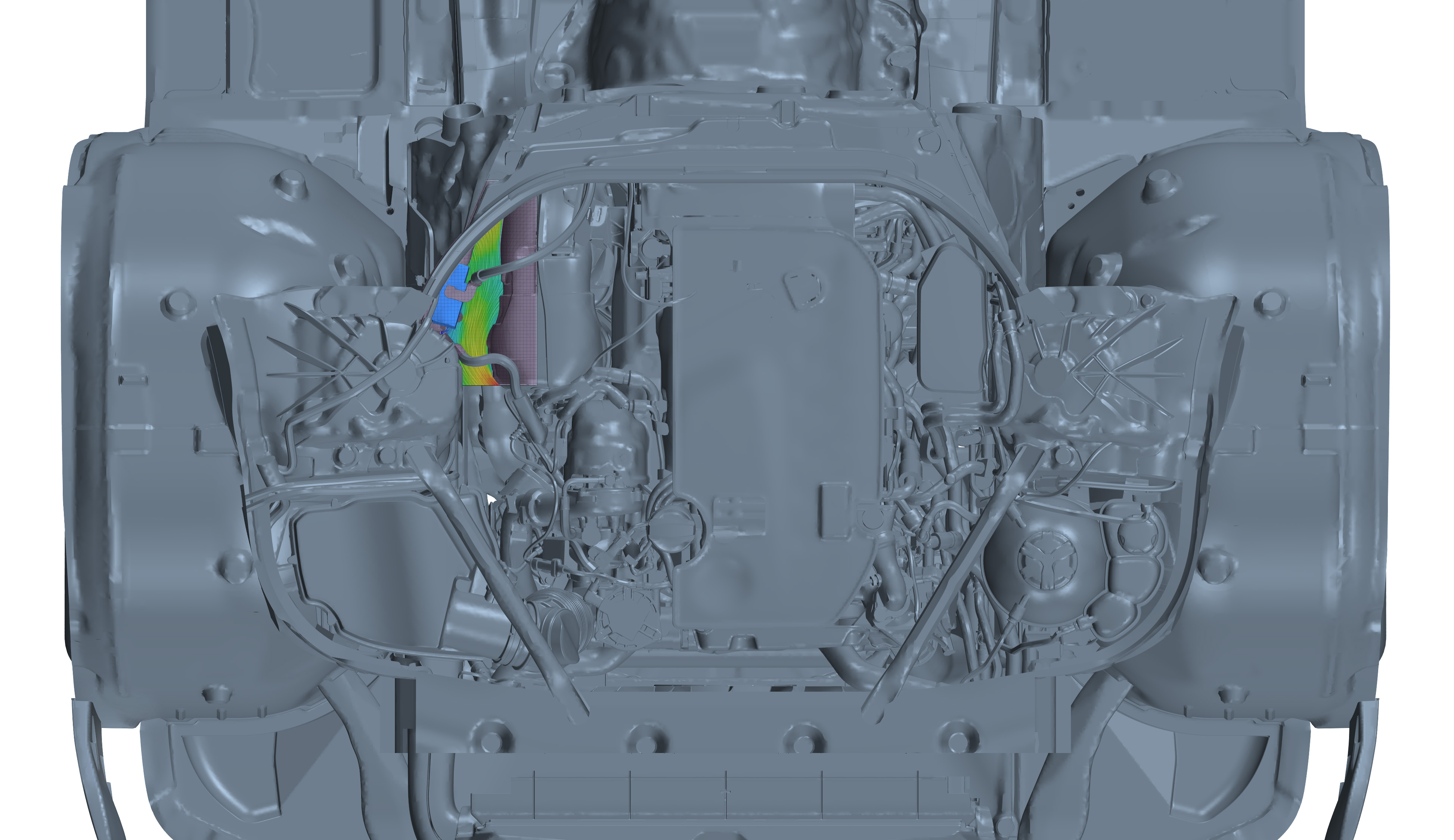}}
		\setlength{\fboxsep}{0pt}
		\setlength{\fboxrule}{0.5pt}
		\fbox{\includegraphics[height=3.0cm]{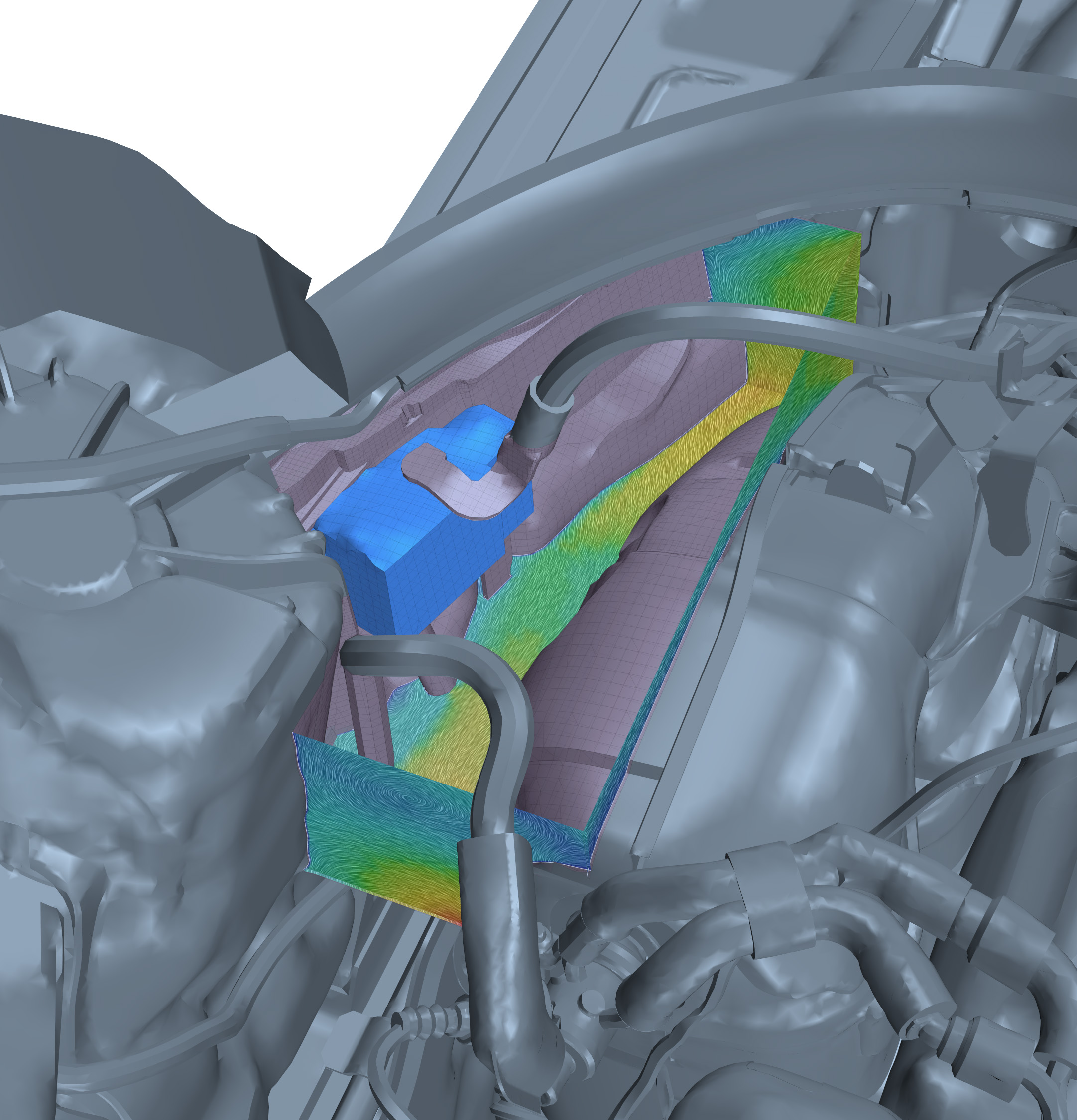}}
		\setlength{\fboxsep}{0pt}
		\setlength{\fboxrule}{0.5pt}
		\fbox{\includegraphics[height=3.0cm]{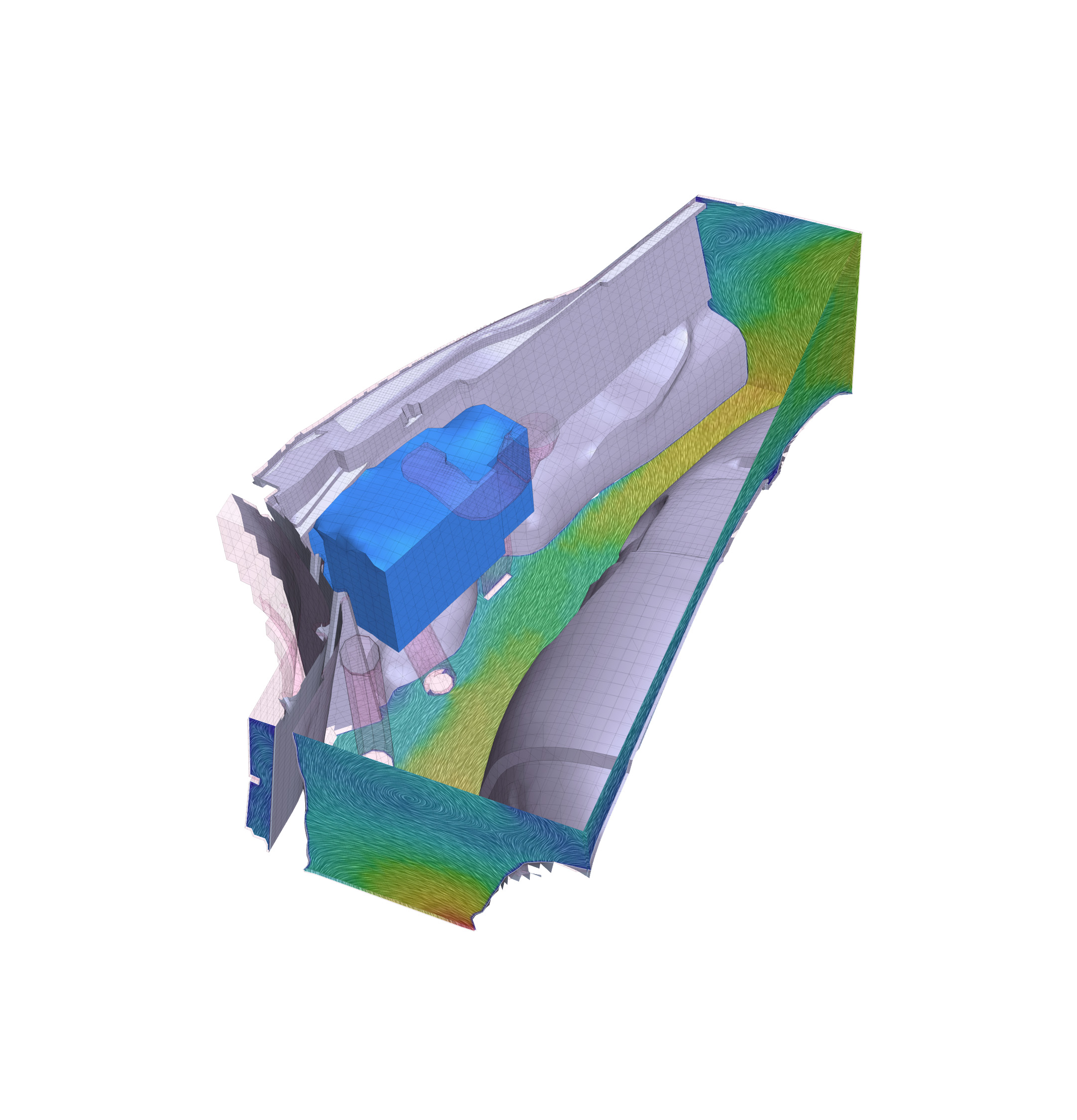}}
		\caption[Configuration of the exhaust system scenario inside a vehicle engine bay]{Configuration 
			of the exhaust system scenario. \textit{Left/Center:} Submodel inside a vehicle engine bay. \textit{Right:} 
			Submodel with cleaned small geometries. Electrical unit modeled as solid is marked in blue.}
		\label{fig:fssp3dp_configuration}
	\end{figure}

	This section presents an exemplary real world application of the proposed PINN approach as well 
	as the suggested thermal coupling based on PINN predicted flow fields. 
	Temperature measurements during vehicle development are expensive as they require prototypes with representative heat sources and airflow. 
	Therefore, digital manufacturing and development processes are integral to reducing the necessity for physical prototypes. 
	To test the PINN's application as part of virtual temperature estimations, a submodel is created from existing full-vehicle simulation results as shown in \reffig{fig:fssp3dp_configuration}. The PINN is used to predict the flow for a thermally coupled simulation of the shown submodel. 
	
	A box, defined as
	\begin{equation*}
		x\in[0.01,0.3]\munit{\,m}\,,\quad y\in[0.275,0.46]\munit{\,m}\,,\quad z\in[0.381,0.55]\munit{\,m}\,,
	\end{equation*}
	is cut from the vehicle engine bay. The r.h.s. of \reffig{fig:fssp3dp_configuration} depicts the 
	resulting submodel, with simplified geometry.
	
	A component of interest is the electrical unit marked in blue, where the maximum surface 
	temperature is to be approximated.
	The flow is predicted using the proposed PINN approach and compared to a 
	CFD reference solution using equal assumptions regarding the governing equations. 
	Moreover, the predicted flow fields are utilized to perform a thermally coupled simulation and investigate the obtained fluid and solid temperature fields.

	The flow and thermal BC at the submodel domain boundaries are extracted from the existing 
	full-vehicle simulation results. 
	As the rear boundary $x=0.3\munit{\,m}$ represents the main outflow, the pressure is prescribed 
	there. At all other cut surfaces the velocity is prescribed while no-slip conditions are applied at all 
	walls.
	The fluid material properties are chosen as $\rho = 1 \munit{\,\frac{kg}{m^3}}$ and $\mu = 
	\expnum{2}{-2} \munit{\Pas}$. This yields a stable, steady and laminar solution for the applied BC.
	
\paragraph*{Remark:} \textit{
		The full-vehicle simulation data is originally generated using Reynolds-averaged Navier-Stokes 
		(RANS) turbulence modeling and material properties corresponding to air. As this cannot be 
		reproduced using the laminar flow equations, the viscosity is chosen such that it roughly matches 
		the effective turbulent viscosity estimated by the turbulence model.
		Although there are already PINN that incorporate the turbulence modeling \cite{ChengEtAl2021} this topic is not investigated in scope of this work.}
	
\begin{figure}[h!]
		\centering
		\includegraphics[width=0.48\textwidth,scale=1.0]{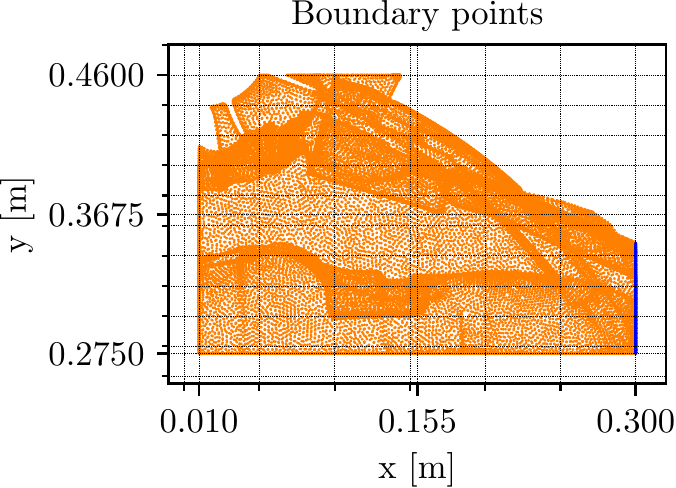}
		%\hfill
		\includegraphics[width=0.48\textwidth,scale=1.0]{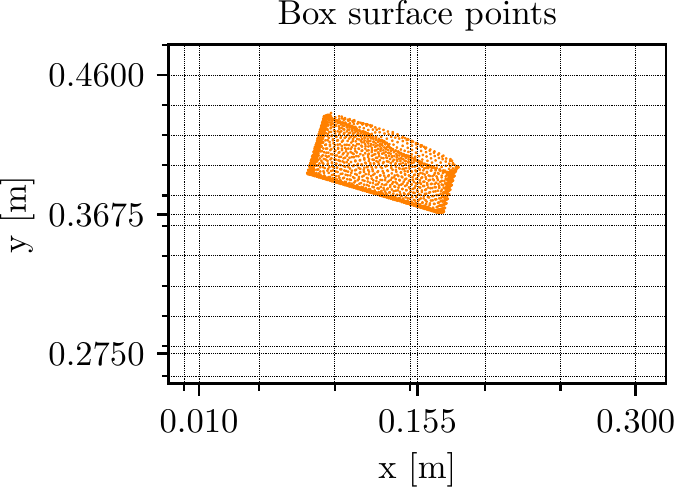}
		\caption[Collocation points of the exhaust system scenario]{Boundary collocation points of the 
			exhaust system scenario. \textit{Left:} All boundary points including Dirichlet set (orange) and Neumann set (blue). \textit{Right}: Only points representing the electrical unit box's surface.}
		\label{fig:meshfssp3d}
	\end{figure}

\begin{figure}[t]
		\centering
		\begin{tabular}{lll}
			\toprule
			CFD & PINN & Difference \\
			\midrule
			\includegraphics[scale=1.0]{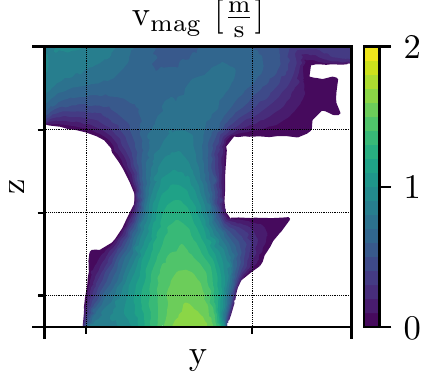} &
			\includegraphics[scale=1.0]{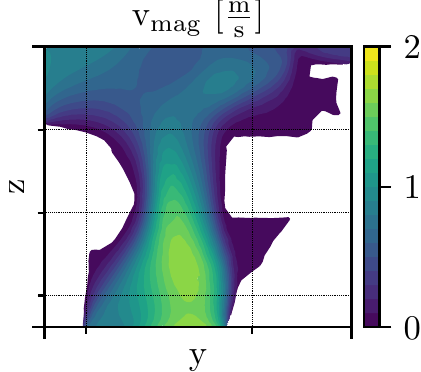} &
			\includegraphics[scale=1.0]{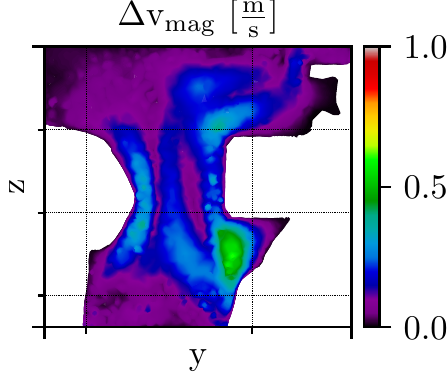} \\
			\includegraphics[scale=1.0]{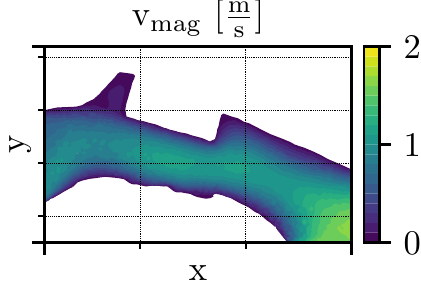} & 
			\includegraphics[scale=1.0]{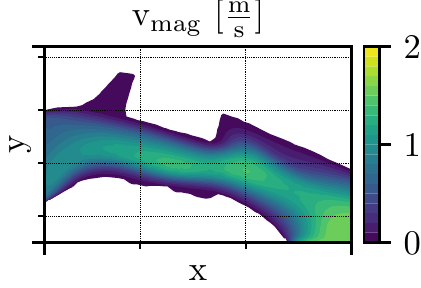} & 
			\includegraphics[scale=1.0]{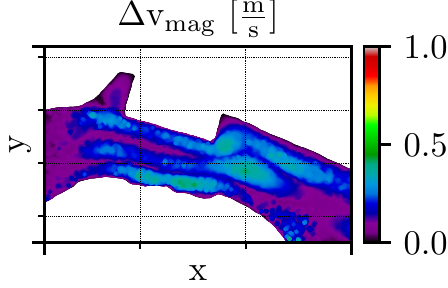} \\
			\includegraphics[scale=1.0]{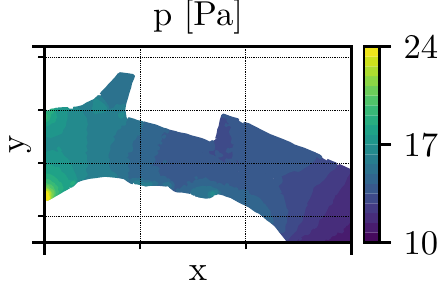} & 
			\includegraphics[scale=1.0]{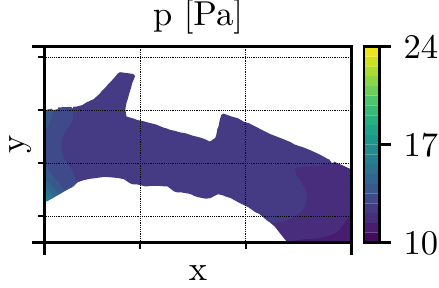} & 
			\includegraphics[scale=1.0]{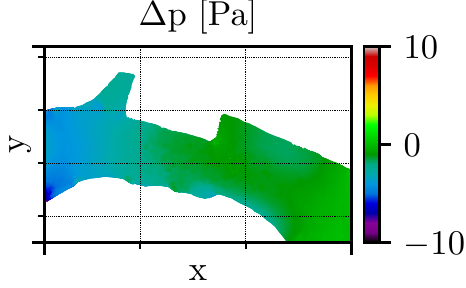} \\
			\bottomrule
		\end{tabular}
		\caption{Three-dimensional flow inside an engine bay scenario. Velocity and pressure cuts at $x=0.12\munit{\,m}$ and $z=0.46\munit{\,m}$. Using the network configuration $n\times m = 10\times 60$,$f_{BC}=10$ and $f_\sigma=1$.}
		\label{fig:fssp235_s1}
	\end{figure}

	The thermal simulation is performed by prescribing the predicted flow solution for the CFD solver. 
	Subsequently, a surface coupled co-simulation of the fluid and a solid domain is performed using Siemens 
	STAR-CCM+ whereas the fluid velocity and pressure are frozen. 
	The solid domain is represented by the electrical unit box. It is modeled using the material 
	properties listed in \reftab{tab:boxMaterial}. The surface directed towards the engine bay's wall is assumed to be adiabatic. 
	To approximate incoming radiation from surrounding geometries missing in the submodel, the 
	environmental radiation temperature is set to $350\munit{\,K}$.
	The domain is resolved by $N_f=128933$ volume collocation points. The Dirichlet and Neumann 
	boundary are represented by $N_D=39278$ and $N_N=1171$ points respectively. The utilized 
	boundary points are illustrated in \reffig{fig:meshfssp3d}. Reference parameters of $L_{ref} = 
	0.16\munit{\,m}$ and $V_{ref} = 3.0\munit{\,\frac{m}{s}}$ are applied. 
	Training is performed using the loss weighting factors $f_{BC} = 100$ and $f_\sigma=1$.
	The obtained flow results are illustrated in \reffig{fig:fssp235_s1}.
	The l.h.s. shows the laminar CFD reference solution. The 
	main flow in positive $x$-direction is visible in both solutions. Flow direction and magnitude are 
	recovered by the PINN.
	The wall boundary layer is approximated significantly thicker than in the 
	reference solution.
	Although the accumulated velocity error at the outlet boundaries is less than $0.1\munit{\,\frac{m}{s}}$, errors as high as $0.3\munit{\,\frac{m}{s}}$ occur in the boundary layer. 
	A vortex formed below the box is not well captured by the PINN.
	This yields a local error of about $0.5\munit{\,\frac{m}{s}}$ but has no significant impact on the main flow.
	The pressure gradient is not well recovered by the PINN.
	A pressure gradient is expected for $x<0.1\munit{\,m}$ but underpredicted by the PINN solution. This yields a maximum pressure error of about $-8\munit{\,Pa}$ at $x=0.01\munit{\,m}$.
	
	\begin{table}[h!]
		\centering
		\caption[Box material properties]{Box material properties.}
		\begin{tabular}{cccc}
			\toprule
			Property & & Value & \\
			\midrule
			Thermal conductivity & $\lambda$ & $0.4$ & $\munit{\,\frac{W}{m\,K}}$ \\
			Surface emissivity & $\epsilon$ & $0.8$ \\
			\bottomrule
		\end{tabular}
		\label{tab:boxMaterial}
	\end{table}
	
\begin{figure}[t]
		\centering
		\begin{tabular}{lll}
			\toprule
			CFD & PINN & Difference \\
			\midrule
			Fluid &&\\
			\includegraphics[scale=1.0]{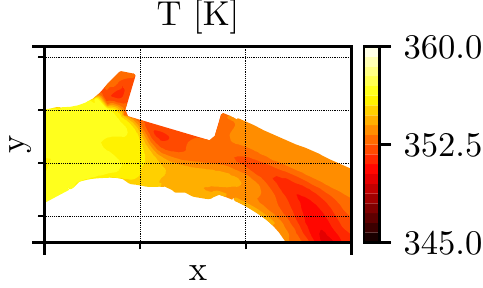} & \includegraphics[scale=1.0]{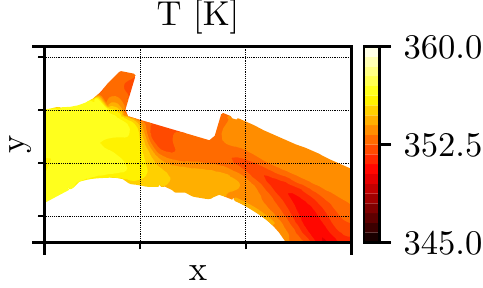} & \includegraphics[scale=1.0]{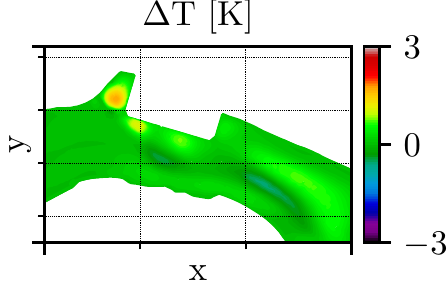} \\
			\midrule
			Box front surface &&\\
			\includegraphics[scale=1.0]{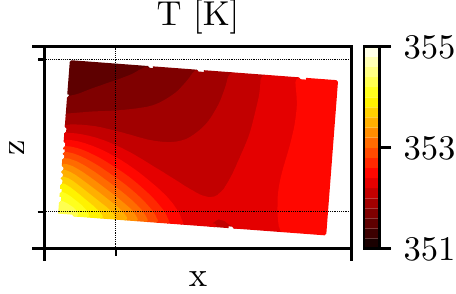} & \includegraphics[scale=1.0]{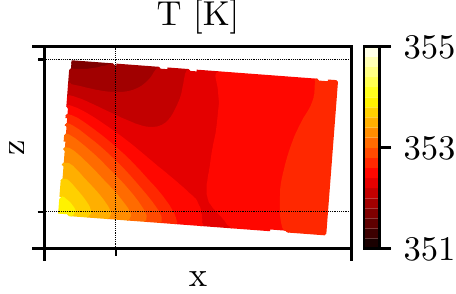} & \includegraphics[scale=1.0]{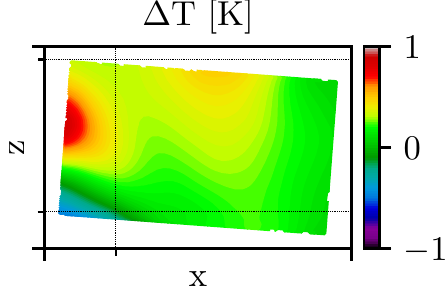} \\
			\bottomrule
		\end{tabular}
		\caption{Fluid temperature field obtained using the predicted flow for an exhaust system scenario flow field with $f_\sigma=1$. Comparison of PINN predicted flow and CFD reference solution. Cuts at $x=0.12\munit{\,m}$ and $z=0.46\munit{\,m}$.}
		\label{fig:fssp35_temps}
	\end{figure}

	The PINN predicted flow field is used to perform a thermally coupled simulation. The resulting fluid 
	and solid temperature field are compared to a coupled CFD solution in \reffig{fig:fssp35_temps}. 
	Overall, the fluid temperature distributions show good agreement. The temperature deviation 
	mostly stays below $\Delta T = \pm 1 \munit{\,K}$. Below and in front of the box are low velocity 
	vortices which are poorly captured by the PINN. Here the maximum temperature deviation of about 
	$2 \munit{\,K}$ is observed. High temperature fluid entering at the front boundary and exiting the 
	domain at the bottom boundary is approximated well. \\
	
	Based on this real world example, we can conclude that the PINN serves it purpose as a surrogate model for flow field estimation in thermally coupled simulations well.
	In general, the temperature deviations recovered on the box's surface are comparatively small and more important the general structured of low and high temperature regions matches visibly, cmp. \reffig{fig:fssp35_temps}.
	Once the PINN is trained, it quickly provides flow field estimate for varying scenarios.
	Particularly in the early stages of vehicle design, the PINN is advantageous over conventionally coupled full-order simulation, as it provides sufficient accuracy while saving a considerable amount of time.
	
	\section{Conclusion}\label{sec:conclusion}
	In scope of this work, we present a novel approach to the forward prediction of parametric flow fields in three spatial dimensions.
	Based on the described dimensionless multivariate scheme, we successively extended the pre-existing PINN method to learn highly nonlinear behavior of the Navier-Stokes equations for varying geometries.
	The presented procedure for training a geometric parameter successfully exploits the PINN's property of trivially known target values for the physics loss.
	The resampling of parameter values during training allows for learning a continuous parameter space based on a single, initial point set.
	This eliminates the necessity for a discrete set of meshes and multiple CFD solutions to model a design parameter variation.\\
	
	We validated the method step-wise for every newly implemented functionality.
	In a first step, we extended the test case of a laminar flow past a two dimensional cylinder used in Rao et Al. \cite{10.1007/978-3-030-22808-8_24} by a third dimension.
	In a second step this test-case was further extended to included parameterization of the cylinder position into the PINN training.
	For both test cases the novel method showed good convergence behavior of the PINN and the PINN predictions were in good accordance with results of full-order simulations.
	In a last validation step we investigated the influence and sensitivity of our predictions with regard to major changes of the dominant direction of flow.
	All test cases used for validation were satisfactorily accurate. 
	Finally, we could successfully apply the novel method in context of this work's original motivation:
	The flow field predicted by the PINN was successfully used in a computation of the surface temperature on temperature-critical vehicle components, directly incorporating the positioning of the component within the PINN.\\
	\\
	We conclude that PINN have great potential to function as surrogate models to approximate complex physical behavior.
	Eventhough the presented implementation is currently limited to a description of the moving domain by logical statements, a pointwise definition of the allowed parameter space would allow for arbitrarily complex geometric variations.
	Further, it should in fact be compatible with any additional input dimension of 
	known effect on the BC such as continuous time modeling.
	Still, looking back on the initial motivation for this work, we already could make use of a PINN as a surrogate model to flow field prediction at varying geometries of temperature sensitive components in vehicle design.
	In this context, PINN already provide huge savings of computational resources and time and enable fast evaluation of new designs.

	\paragraph{Acknowledgments}
	The authors gratefully acknowledge the support and computing resources provided by the Bayerische Motoren Werke AG.\\
	\\
	Funded by the Deutsche Forschungsgemeinschaft (DFG, German
	Research Foundation) under Germany’s Excellence Strategy – EXC-
	2023 Internet of Production – 390621612. 
	
	
	\bibliographystyle{ieeetr}  
	\bibliography{MyBib}
	
\end{document}